\documentclass[twocolumn]{dslabarticle}

\usepackage{amsmath}
\usepackage{amsthm}
\usepackage{amssymb}

\usepackage{url}
\usepackage{caption}
\usepackage{pifont}
\usepackage[T1]{fontenc}
\usepackage{amsfonts}
\usepackage{nicefrac}
\usepackage{microtype}

\usepackage{xparse}

\usepackage[utf8]{inputenc}
\usepackage{hyperref}
\usepackage{xurl}

\usepackage{graphicx, color}
\usepackage{dblfloatfix}

\usepackage{booktabs}
\usepackage{longtable}

\definecolor{shadecolor}{RGB}{200,200,200}
\NewDocumentCommand{\codeword}{m}{%
\texttt{#1}
}

\title{AnySleep: a channel-agnostic deep learning system for high-resolution sleep staging in multi-center cohorts}

\author[1,2,3]{Niklas Grieger}
\author[1,3]{Jannik Raskob}
\author[2]{Siamak Mehrkanoon}
\author[1,3]{Stephan Bialonski}

\affiliation[1]{Department of Medical Engineering and Technomathematics, FH Aachen University of Applied Sciences, Jülich, Germany}
\affiliation[2]{Department of Information and Computing Sciences, Utrecht University, Utrecht, The Netherlands}
\affiliation[3]{Institute for Data-Driven Technologies, FH Aachen University of Applied Sciences, Jülich, Germany}

\abstract{
Sleep is essential for health, yet studying its dynamics requires manual sleep staging, a labor-intensive step in research and clinical care.
Across centers, polysomnography (PSG) recordings are traditionally scored in 30-s epochs for pragmatic, not physiological, reasons and vary in electrode count, montage, and subject characteristics.
These constraints challenge harmonized multi-center studies and the discovery of robust biomarkers on shorter timescales.
We present AnySleep, a deep neural network that scores sleep from any electroencephalography (EEG) or electrooculography (EOG) data at adjustable temporal resolutions.
We trained and validated the model on over 20,000 overnight recordings ($>$ 200,000~h of EEG and EOG) from 28 datasets across multiple clinics to promote robust generalization across sites.
The model attains state-of-the-art performance and surpasses or equals established baselines at 30-s epochs.
Performance improves with more channels, yet remains strong when EOG is absent, or only EOG or single EEG derivations (frontal, central, or occipital) are available.
On sub-30-s timescales, the model captures short wake intrusions consistent with arousals and improves the prediction of pathophysiological conditions (obstructive sleep apnea, narcolepsy type 1, insomnia) over 30-s scoring.
We make the model publicly available to facilitate large-scale studies with heterogeneous electrode setups and accelerate biomarker discovery in sleep.
}

\metadata[Model and code]{\url{https://github.com/dslaborg/AnySleep}}
\correspondence{N.G. (\email{grieger@fh-aachen.de}), S.B. (\email{bialonski@fh-aachen.de})}
\metadata[Published version]{npj Digital Medicine 9, 829 (2026). \url{https://doi.org/10.1038/s41746-026-03406-7}}

\def\USleep{\mbox{U-Sleep}}

\renewcommand{\addcontentsline}[3]{}

\begin{document}

\maketitle

\section{Introduction}

Sleep carries diagnostic and prognostic value across a wide range of conditions, from sleep disorders to cardiometabolic, psychiatric, and neurodegenerative diseases.
In clinical practice and research, extracting this information usually requires overnight polysomnography (PSG) and expert annotations (sleep staging), which is work-intensive, costly, and subject to significant inter-rater variability~\cite{Cesari2021b,Lee2022,Nikkonen2023}.
Moreover, sleep dynamics have traditionally been analyzed based on 30-s epochs, a convention originating from the practical constraints of manual annotation on paper strips rather than from any underlying physiological rationale~\cite{Loomis1936,Decat2022}.
This approach has long served as the foundation for sleep research, yet sleep unfolds at temporal resolutions much finer than can be captured by 30-s windows.
This is particularly evident in gradual sleep state transitions, which may pass through short intermediate ``substages''~\cite{Lambert2023,Decat2022}, or in brief disruptions caused by micro-sleep or micro-arousals.
The latter occur on the scale of seconds and play a critical role in various sleep disorders, including REM sleep behavior disorder (RBD), obstructive sleep apnea (OSA), and insomnia~\cite{Korkalainen2021,BrinkKjaer2020,Malafeev2021,Cesari2021,Younes2015,Moul2007}.

Large-scale studies with shorter timescale annotations would, therefore, be valuable for gaining a deeper understanding of sleep dynamics and discovering novel biomarkers.
Yet, despite the availability of large amounts of raw PSG data, conducting large-scale studies based on expert annotations is practically infeasible, as the time and effort required for manual scoring increases substantially with the frequency of annotations.
At the same time, empirical studies have demonstrated that shorter annotations lead to increased inter-rater variability~\cite{Follin2025b,Moul2007,Tautan2022}, although it has been hypothesized that shorter epochs could reduce disagreements by decreasing the number of ambiguous transition epochs~\cite{Follin2025}.

Automated sleep staging models based on machine learning offer a potential solution to these limitations.
These models can quickly and cost-effectively provide sleep annotations at high resolution even when trained with conventional 30-s labels (e.g., via shorter input windows~\cite{Stephansen2018,Tautan2022,Koch2018}, high-resolution feature maps~\cite{Olesen2020}, segmentation architectures~\cite{Perslev2021,Zan2023}, or sliding-window methods that generate overlapping predictions per 30-s window~\cite{Korkalainen2021,Krauss2021}).
Yet, adapting these models for large-scale studies requires a combination of robust generalization across cohorts and clinics, as well as flexible handling of heterogeneous montages.
Furthermore, models should be able to provide sleep stage predictions at configurable resolutions, which is essential for studying sleep dynamics across a wide range of timescales.
While recent work has made progress on these aspects individually or in pairs, including generalization~\cite{Olesen2020,Guillot2021,Perslev2021,Vallat2021,Hanna2023}, robustness to varying channel configurations~\cite{Guillot2021,Shi2024}, and flexible high-resolution predictions~\cite{Perslev2021}, a single approach combining all three capabilities remains an open problem.

Among recent models, \USleep{} comes closest to meeting these requirements, providing sleep stage predictions at adjustable temporal resolutions of up to 128~Hz and showing robust generalization across datasets~\cite{Perslev2021}.
However, \USleep{}'s practical utility is limited by its deliberately fixed input modality and channel requirements (one EEG and one EOG channel in the original version, single channels in later developments~\cite{Rossi2025b}), which does not natively support integration of heterogeneous multi-channel configurations encountered in large-scale studies.
Restricting inputs to one or two channels also limits the spatial resolution across the scalp, which does not reflect the recommendations of the American Academy of Sleep Medicine (AASM) to use at least three EEG channels placed at frontal, central, and occipital regions of the scalp, as well as EOG and EMG channels for sleep scoring~\cite{Berry2020}.
Although these recommendations were developed for human scoring, empirical evidence suggests that automated systems likewise benefit from access to additional channels~\cite{Chambon2018,Phan2019,Guillot2021}.
To enable \USleep{} to handle more than two channels, it was proposed to evaluate recordings multiple times with different channel combinations and to aggregate the resulting predictions by majority vote~\cite{Perslev2021}, a post-hoc strategy that can serve as a pragmatic workaround but lacks an explicit mechanism for learning complex cross-channel relationships and scales poorly as channel counts increase.

In this work, we introduce AnySleep, a deep neural network that can dynamically combine any available EEG or EOG channels to score sleep at flexible temporal resolutions.
We trained and validated the model on 20,589 overnight recordings from 28 datasets spanning multiple centers, recording setups, and patient populations, and assessed generalization on datasets from studies not used during training.
We evaluated AnySleep at conventional 30-s epochs across diverse channel configurations, including cases with missing EOG or EEG channels, and examined how performance depended on the number of available channels.
At shorter timescales, we studied whether AnySleep's high-resolution predictions could represent short sleep events, such as arousals, and whether they provide micro-architectural information useful for distinguishing between healthy controls and patients with obstructive sleep apnea (OSA), narcolepsy type 1 (NT1), or insomnia.
AnySleep's ability to characterize sleep dynamics at short timescales and across heterogeneous montages makes it a promising tool with the potential to accelerate the discovery and validation of novel biomarkers.

\section{Results}

\subsection{Datasets and Model Training}

We trained and evaluated AnySleep on an extensive collection of 28 datasets comprising 20,589 overnight recordings ($\approx$~200,000~h of EEG/EOG data).
These datasets cover a wide range of recording conditions, including different clinics, recording setups, patient populations, and experts.
We divided the datasets into two groups: an in-distribution group of 13 datasets and a hold-out group of 15 datasets (see Section~\ref{ssec:data}).
While the in-distribution group served as the basis for model training and validation, the hold-out group was solely used for testing and split into subsets for traditional 30-s sleep staging (8 datasets) and high-resolution analysis (7 datasets).
By isolating the hold-out group from the training process, we aimed to obtain a realistic measure of our model's ability to generalize to new datasets from other studies and clinics.
To enable comparisons between AnySleep's and \USleep{}'s scoring performance, our data splits closely followed those used by \USleep{}~\cite{Perslev2021}.

The design of AnySleep blends two architectural concepts: a U-Net-inspired encoder-decoder architecture~\cite{Ronneberger2015} that allows for high-frequency sleep staging, and channel-attention modules that enable the model to handle variable numbers and choices of EEG and EOG channels (see Section~\ref{ssec:model}).
In brief, each input channel is first processed by successive encoder blocks to yield channel-specific feature maps at increasing levels of abstraction.
Channel-attention modules then combine a variable number of these channel-specific feature maps into cross-channel representations based on learnable attention weights that specify each channel's relevance to the sleep staging task.
The cross-channel representations are passed to the decoder branch of the architecture, at which end a segment classifier produces sleep stage predictions at configurable frequencies of up to 128~Hz, corresponding to a minimum temporal resolution of about 0.008~s per predicted sleep stage.
To encourage robustness to heterogeneous montages, we trained AnySleep while randomly varying both the number and type of input channels (see Section~\ref{ssec:training}).

\subsection{Robustness to Channel Configurations}
\label{ssec:channel-robustness}

\begin{figure*}[t]
  \centering
  \includegraphics[width=\linewidth]{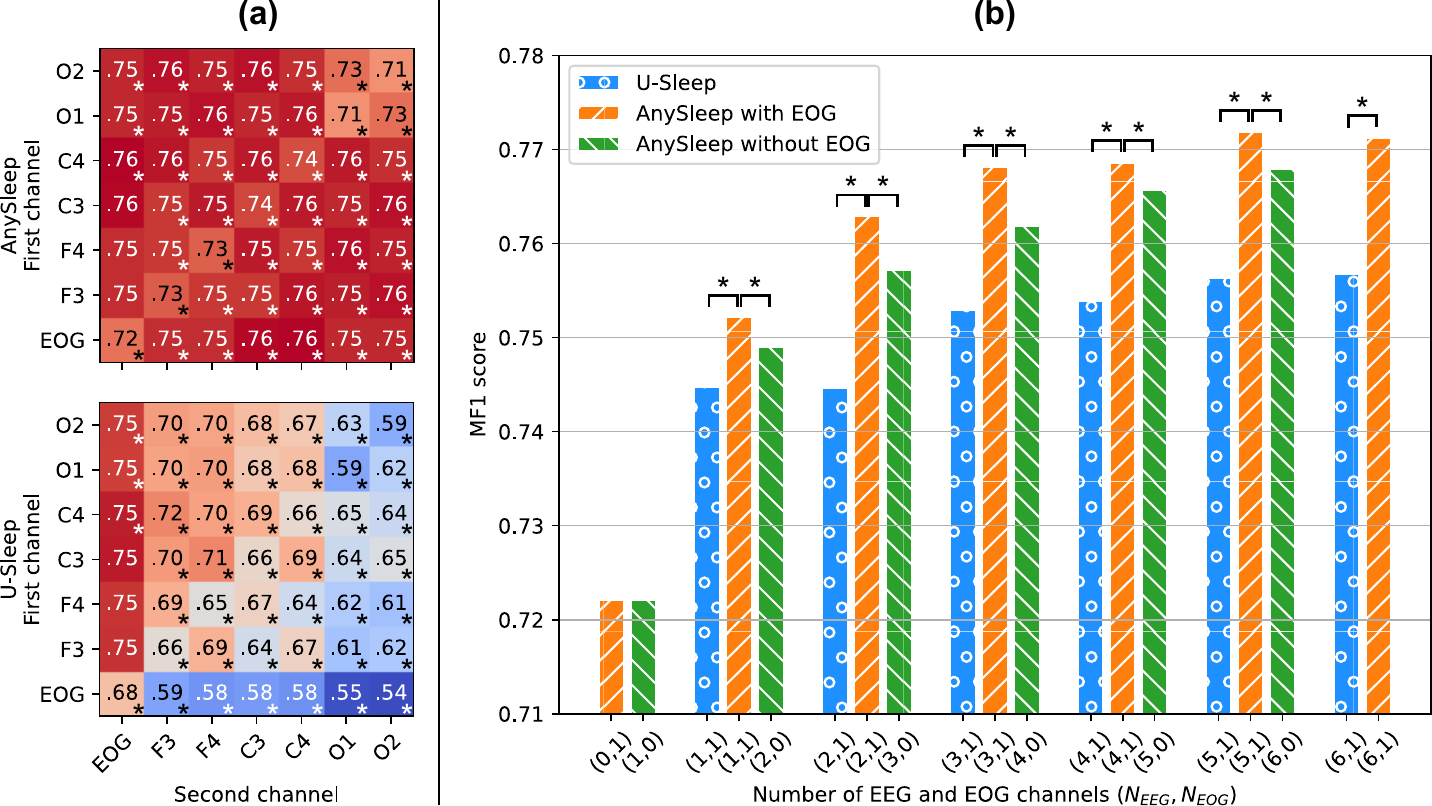}
  \caption{
    Robustness of AnySleep and \USleep{} to variations in channel type, order, and number as measured by recording-wise macro F1 scores (higher values indicate better performance).
    \textbf{(a)} Performance of AnySleep (upper matrix) and \USleep{} (lower matrix) for two-channel input permutations using EOG and frontal (F3/F4), central (C3/C4), and occipital (O1/O2) EEG channels; diagonal entries correspond to single-channel configurations.
    Evaluations used all test recordings in the reference montage that contained all investigated channels (563 recordings; see Supplementary Table~\ref{sup-tab:channels} for detailed channel overview).
    \textbf{(b)} Performance for a varying number of EEG ($N_\text{EEG}$) and EOG ($N_\text{EOG}$) channels, grouped by the total number of channels.
    Evaluations were restricted to recordings with at least one EOG and six EEG channels from the test set (643 recordings).
    For each channel-count condition, we randomly sampled 5000 recordings with replacement and evaluated each on a randomly chosen subset of channels of that size (without replacement).
    Missing bars reflect \USleep{}'s multi-channel requirement and the experiment's restriction to six EEG channels.
    In both panels, evaluations were repeated for three independent training runs, and we report the average scores over recordings and runs.
    Stars (*) indicate significant ($p<0.01$, one-sided t-test) differences in scores between AnySleep and \USleep{} (see \textbf{(a, b)}), or between AnySleep with and without EOG (see \textbf{(b)}).
  }
  \label{fig:channel_comb_matrix}
\end{figure*}

We assessed AnySleep's dependence on channel type, spatial location, and input order by testing the model with different single- and two-channel configurations.
Specifically, we evaluated test recordings on all possible single- or two-channel permutations created from the following set of channels: EOG (left or right), F3, F4, C3, C4, O1, and O2 (e.g., F3, F3 \& EOG1, EOG1 \& F3).
For each recording and channel permutation, we predicted 30-s sleep stages and compared these predictions with expert annotations to obtain macro F1 (MF1) scores as measures of model performance.
We repeated the same analysis for the \USleep{} model, duplicating the input channel in configurations where AnySleep was evaluated with a single channel to account for \USleep{}'s inability to handle single-channel inputs.

AnySleep showed minimal sensitivity to channel type, spatial location, or input order (see Figure~\ref{fig:channel_comb_matrix}a).
Across all two-channel configurations, including those without EOG, macro F1 scores lay in a narrow range (0.726--0.760).
Performance remained high under single-channel conditions: the lowest score of 0.710, obtained when only one occipital EEG derivation was provided, was only slightly below the maximum score of 0.760 observed for the best two-channel combinations.
In comparison, \USleep{} was originally designed to receive exactly one EEG and one EOG channel, and performance decreased substantially when deviating from this design choice by swapping the order of EEG and EOG channels (average macro F1 decrease of 0.160--0.203; see Figure~\ref{fig:channel_comb_matrix}a).
We observed a similar, albeit less severe, decline in performance when replacing the EOG channel with a second EEG channel, particularly a frontal one.
This suggests that \USleep{} can partially exploit eye-movement information embedded in frontal EEG that is less pronounced at more posterior sites.

Next, we investigated how performance depended on the number of input channels (see Figure~\ref{fig:channel_comb_matrix}b).
We evaluated AnySleep and \USleep{} on test recordings with random channel subsets containing between one and seven channels, with at most one EOG channel included.
Given a recording and a channel subset, we evaluated AnySleep with a single forward pass, while \USleep{} was evaluated on all possible (EOG, EEG) channel pairs ($N_{\text{EEG channels}} \cdot N_{\text{EOG channels}}$ runs) with subsequent majority voting~\cite{Perslev2021}.
AnySleep's macro F1 increased with the number of available EEG channels, reaching 0.771 when six EEG and one EOG channels were provided.
This trend persisted at even larger channel counts, as observed on the MASS datasets (up to 23 channels) and an additional high-density dataset (85 channels; see Supplementary Table~\ref{sup-tab:high-density}), with no substantial differences between channels represented in the training datasets and those not represented (see Supplementary Figure~\ref{sup-fig:nchannels-mass}).
When the EOG channel was omitted and replaced with an additional EEG derivation, performance decreased slightly but consistently, suggesting that an additional modality (EOG) provides more complementary information than adding another EEG channel.
Across all tested channel numbers, AnySleep achieved higher macro F1 scores than \USleep{}, which seemed to benefit less from additional EEG channels, likely reflecting AnySleep's more flexible and dynamic handling of multi-channel input.

\subsection{High-Frequency Sleep Staging Capabilities}
\label{ssec:high-freq-results}

\begin{figure*}[ht!]
  \centering
  \includegraphics[width=\linewidth]{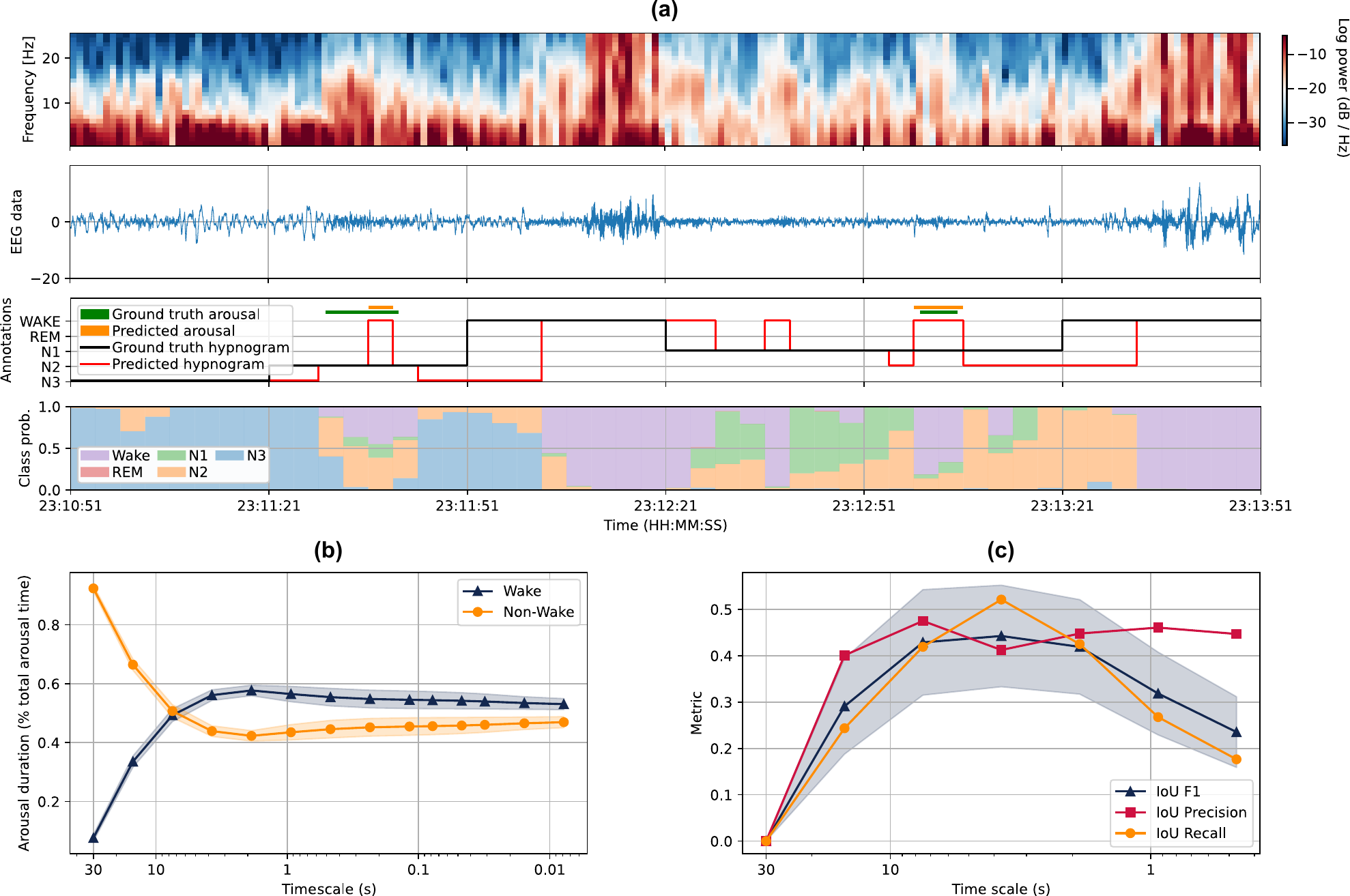}
  \caption{
    Representation of short arousals in AnySleep's high-frequency sleep stage predictions on held-out MASS C1 and C3 data.
    \textbf{(a)} Three-minute EEG segment from MASS~C1.
    The top panel shows a time-frequency representation (spectrogram), the middle panel the raw EEG trace (C4-CLE, CLE = computed linked-ear~\cite{OReilly2014}), and the bottom panel the corresponding annotations (black line: expert-scored 30-s sleep stages; red line: AnySleep predictions at 3.75-s resolution; colored areas: class probabilities for Wake, REM, N1, N2, N3; green bars: expert-annotated arousals; orange bars: arousals derived from high-resolution Wake predictions; see Section~\ref{ssec:high-freq-results}).
    \textbf{(b)} Proportion of total expert-annotated arousal time in MASS C1 and C3 that overlaps with intervals predicted as Wake (dark blue) or non-Wake (orange) by AnySleep, as a function of the temporal resolution of sleep stage predictions.
    Curves show the mean across three independently trained models; shaded areas indicate the standard deviation across models.
    \textbf{(c)} Arousal detection performance in MASS C1 and C3 at different sleep stage resolutions, quantified using intersection-over-union (IoU) precision, recall, and F1 score (0 = no agreement, 1 = perfect agreement) between predicted and expert arousals.
    A predicted and an expert arousal were counted as matching if their temporal overlap covered at least 20\% of their combined duration.
    Scores were computed per subject for each of three models, then averaged across subjects and models; the shaded area shows the corresponding standard deviation of the IoU F1 score.
  }
  \label{fig:arousals}
\end{figure*}

Because there is currently no widely adopted standard for expert sleep staging at sub-30-s timescales~\cite{Follin2025b,Moul2007}, we assessed AnySleep's high-resolution predictions indirectly, asking whether they encode physiologically meaningful information beyond conventional 30-s epochs.
Visual inspections suggested that the high-frequency predictions captured transitions between sleep states more accurately than 30-s epochs (e.g., Wake transitions in Figure~\ref{fig:arousals}a at around 23:12:05 and 23:13:30).
This was especially evident for arousals, often described as short awakenings~\cite{BrinkKjaer2020,Hermans2022,Younes2015}, which we investigated by comparing expert-annotated arousals in the held-out MASS C1 and C3 test datasets with Wake predictions of AnySleep at different temporal resolutions.
With conventional 30-s predictions, only 7.7\% of the total duration of expert-annotated arousal time overlapped with Wake stages (see Figure~\ref{fig:arousals}b), highlighting the difficulty of representing short events like arousals in traditional sleep staging.
This overlap increased with the temporal resolution of the sleep stage predictions, peaking at 57.7\% at a timescale of around 2~s, and then decreased slightly at even finer timescales (53.1\% at around 0.05~s).

To test whether AnySleep indeed learned to represent arousals as short Wake events, we derived candidate arousals in MASS C1 and C3 by identifying contiguous Wake segments of 3--15~s in the model's predictions (see Figure~\ref{fig:arousals}a; see Section~\ref{ssec:high-freq} for details).
We then compared candidate and expert-annotated arousals using intersection-over-union (IoU) precision, recall, and F1 scores, where 0 indicates no agreement and 1 perfect agreement.
Across temporal resolutions, performance was highest for timescales between 2--8~s, with a maximum IoU precision of 0.475, IoU recall of 0.521, and IoU F1 of 0.442 (see Figure~\ref{fig:arousals}c), corresponding to approximately 52.1\% of expert-annotated arousals being detected and 47.5\% of predicted arousals overlapping with an expert annotation.
As expected, IoU scores declined for sleep stage predictions at timescales longer than 8~s, consistent with short arousals typically being missed in 30-s sleep stages.
At fine resolutions below 2~s, IoU F1 also decreased, suggesting an increasing level of noise in high-frequency sleep stages.

\begin{figure*}[t]
  \centering
  \includegraphics[width=\linewidth]{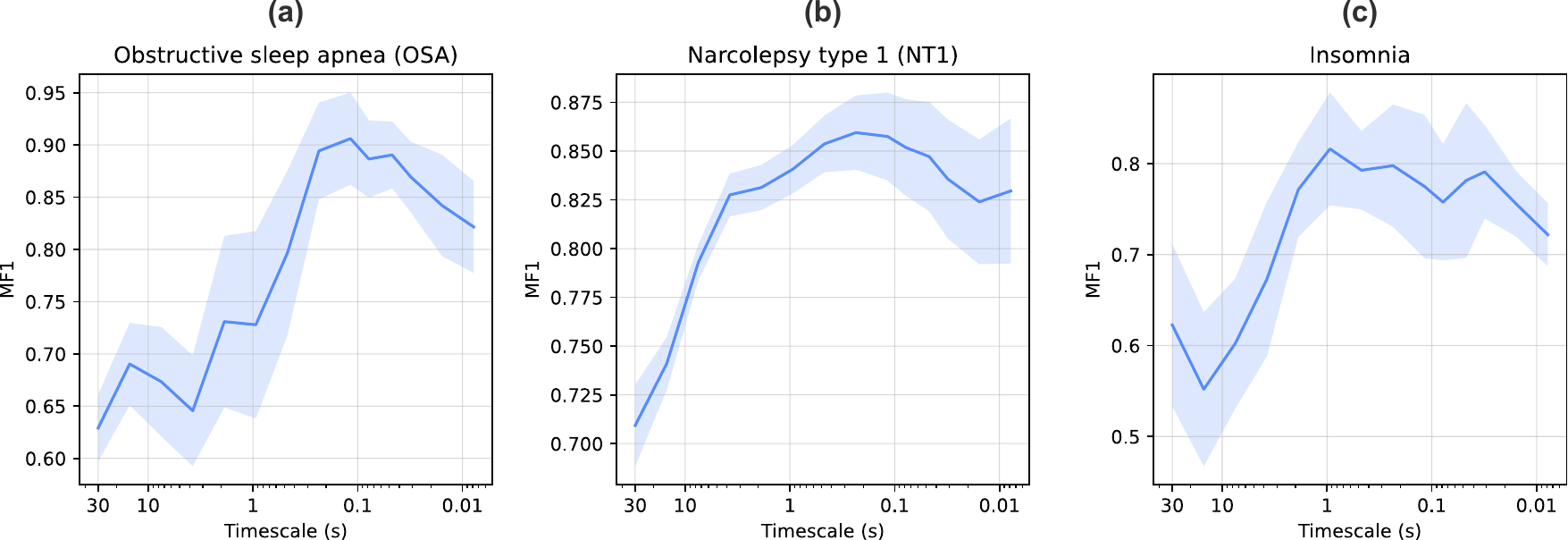}
  \caption{
    Prediction of sleep disorders from triplet features derived from AnySleep's high-frequency sleep stages at different timescales on held-out test datasets.
    \textbf{(a) -- (c)} show macro F1 scores (MF1s) of random forest (RF) classifiers predicting the presence (yes vs no) of obstructive sleep apnea (25 healthy controls from DODH and 55 patients from DODO), narcolepsy type 1 (395 healthy controls and 260 patients from the CNC, DHC, FHC, IHC, KHC, and SSC datasets), and insomnia (9 insomnia patients and 16 healthy controls from the CAP dataset), respectively.
    For each timescale and disorder in (a) -- (c), 50 RF models were trained on features derived from each of three independent AnySleep training runs (150 RF models in total); the blue line and blue shaded area show mean score and standard deviation across these models.
  }
  \label{fig:high-freq}
\end{figure*}

We next investigated whether high-frequency sleep stages could be used to predict subject-level pathophysiological characteristics, such as the presence of obstructive sleep apnea (OSA), narcolepsy type 1 (NT1), and insomnia.
We hypothesized that these characteristics would be reflected in the frequency of rapid transitions between sleep stages, since OSA, NT1, and insomnia have been associated with sleep fragmentation and altered sleep-stage transition dynamics~\cite{Bonnet2003,Korkalainen2021,Christensen2015,Conte2023,Baglioni2014}.
Following the approach of Perslev~et~al.~\cite{Perslev2021}, we quantified the temporal regularity of sleep by using ``triplet features,'' defined as counts of sleep stage triplets $(s_i, s_{i+1}, s_{i+2})$ with $s_i \neq s_{i+1}$ and $s_{i+1} \neq s_{i+2}$.
Varying the resolution of the underlying sleep stage predictions, we calculated the absolute number of these triplets for the subjects of different cohorts and then trained random forest (RF) classifiers to distinguish patients from healthy controls (see Section~\ref{ssec:high-freq}).
Specifically, we trained separate models to distinguish healthy controls from: (i) patients with obstructive sleep apnea (OSA) in the DODO and DODH datasets, (ii) patients with narcolepsy type 1 (NT1) in the CNC, DHC, FHC, IHC, KHC, and SSC datasets, and (iii) patients with insomnia in the CAP dataset.

Compared with conventional 30-s staging, AnySleep's high-frequency sleep stage predictions allowed the RF classifiers to better discriminate between patients and healthy controls across all three disorders (see Figure~\ref{fig:high-freq}).
The best performances were achieved for timescales between 0.1--1.0~s, with MF1 scores of 0.91, 0.86, and 0.82 for the classifications of OSA, NT1, and insomnia, respectively.
Consistent with our findings for arousal detection, performance declined slightly when the temporal resolution was further increased.

To better understand which sleep-transition patterns were most informative for distinguishing patients from healthy controls, we examined the triplets most consistently selected by the classifiers at the temporal resolution that performed best for each disorder (see Section~\ref{ssec:high-freq}).
For OSA, the triplets N2$\rightarrow$Wake$\rightarrow$N2 and REM$\rightarrow$N2$\rightarrow$REM were among the most influential, consistent with pronounced sleep fragmentation and the general vulnerability of non-REM and REM sleep associated with the disorder~\cite{Waechter2019,Bianchi2010}.
We also found that the triplet N2$\rightarrow$N3$\rightarrow$N2 was often selected, which may reflect altered NREM sleep dynamics contributing to the distinction between OSA patients and healthy controls.
For NT1, transitions between REM and wakefulness (e.g., Wake$\rightarrow$N1$\rightarrow$REM, N2$\rightarrow$Wake$\rightarrow$REM, REM$\rightarrow$Wake$\rightarrow$N1, REM$\rightarrow$Wake$\rightarrow$REM) were highly influential, consistent with hypothesized nocturnal SOREMPs and generally altered REM sleep~\cite{Christensen2015,Liu2015}.
Additionally, the influence of the triplet Wake$\rightarrow$N3$\rightarrow$Wake may reflect broader sleep instability in NT1.
For insomnia, we found triplets such as Wake$\rightarrow$N2$\rightarrow$Wake, N1$\rightarrow$Wake$\rightarrow$N2, and N2$\rightarrow$Wake$\rightarrow$N2 were most informative, in line with previously reported NREM instability in insomnia patients~\cite{Hermans2021,Wei2017}.

\subsection{Channel-Attention Patterns}

\begin{figure}[h]
  \centering
  \includegraphics[width=\linewidth]{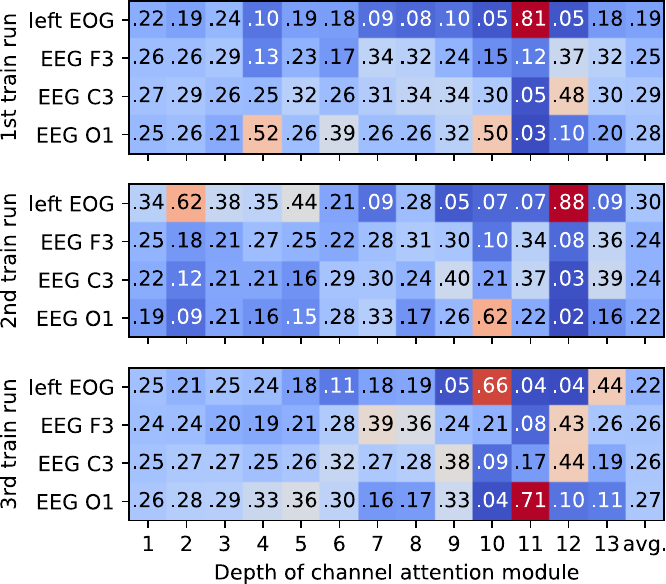}
  \caption{
    Attention weights assigned to the left EOG and EEG F3, C3, and O1 channels by AnySleep's channel-attention modules (columns 1--13).
    Heatmaps show average weights over test recordings containing all four channels in a referential montage.
    The three panels correspond to independent training runs; the rightmost column (``avg.'') shows the mean weight per channel across all modules.
  }
  \label{fig:attention-weights}
\end{figure}

AnySleep's flexibility with respect to varying input channel configurations is supported by channel-attention modules, which learned to assign an attention weight to each available channel.
We analyzed these weights to characterize the model's channel-selection strategy.
For this analysis, we considered test recordings that contained the left EOG and F3, C3, O1 channels in referential montage to maximize the amount of included recordings (614 recordings in total).
Each recording was passed through the trained model, and the attention weights assigned to each channel were extracted from all 13 channel-attention modules located at different depths of the U-Net-inspired architecture (see Section~\ref{ssec:model}).
For every channel and module, we then averaged the attention weights over all evaluated recordings.

We observed varying patterns between the attention modules, with the model focusing on different channels and modalities at different depths (i.e., feature abstraction levels; see Figure~\ref{fig:attention-weights}).
Interestingly, attention patterns differed between training runs, making it unlikely that channel preferences at a given depth are rigidly determined by the receptive field or the characteristic timescale at that depth (for example, targeting a specific frequency band).
This variability is consistent with the stochastic nature of model training, including random weight initialization and stochastic sampling of training epochs and input channels (see Section~\ref{ssec:training}).
Despite the differences in attention patterns, the three training runs achieved similar test performance (see Table~\ref{tab:architectures}), indicating that comparable predictive performance can arise from different channel-weighting preferences.
Nevertheless, two trends were consistent across runs.
First, deeper modules sometimes concentrated most of their weight on a single channel, with average weights of up to 88\%, demonstrating that the model can reliably identify particularly informative modalities or brain regions across recordings.
Second, averaging attention across all modules produced similar mean weights for all four channels, which suggests that AnySleep integrates information from all available EEG and EOG channels rather than relying on a single modality or channel to achieve optimal performance.

\subsection{Impact of Channel-Attention Placement}
\label{ssec:fusion_placement}

\begin{table*}[t]
  \centering
  \begin{tabular}{llrr|c|ccc}
              & Dataset   & $N_{\text{Rec}}$ & $N_{\text{Ch}}$ & \USleep{}            & early fusion         & AnySleep             & late fusion          \\
    \toprule
    In-Dist.  & abc       & 20               & 8               & 0.76 (.009)          & 0.77 (.002)          & \textbf{0.80} (.006) & 0.78 (.006)          \\
    test sets & ccshs     & 78               & 4               & 0.86 (.003)          & 0.85 (.003)          & \textbf{0.87} (.001) & 0.86 (.002)          \\
              & cfs       & 92               & 4               & 0.82 (.004)          & 0.81 (.002)          & \textbf{0.83} (.001) & 0.82 (.002)          \\
              & chat      & 128              & 10              & 0.84 (.007)          & 0.82 (.007)          & \textbf{0.86} (.001) & 0.85 (.002)          \\
              & dcsm      & 39               & 8               & \textbf{0.81} (.005) & 0.80 (.002)          & \textbf{0.81} (.005) & 0.80 (.011)          \\
              & hpap      & 36               & 8               & 0.77 (.002)          & 0.74 (.005)          & \textbf{0.79} (.006) & 0.77 (.001)          \\
              & mesa      & 100              & 5               & 0.78 (.006)          & 0.76 (.007)          & \textbf{0.80} (.001) & 0.79 (.002)          \\
              & mros      & 134              & 4               & 0.76 (.007)          & 0.75 (.003)          & \textbf{0.78} (.001) & 0.77 (.002)          \\
              & phys      & 100              & 7               & \textbf{0.79} (.005) & 0.76 (.004)          & \textbf{0.79} (.002) & 0.78 (.005)          \\
              & sedf-sc   & 23               & 3               & 0.80 (.003)          & 0.80 (.007)          & \textbf{0.81} (.004) & \textbf{0.81} (.002) \\
              & sedf-st   & 8                & 3               & 0.77 (.005)          & 0.76 (.011)          & 0.77 (.004)          & \textbf{0.79} (.003) \\
              & shhs      & 140              & 4               & 0.79 (.005)          & 0.78 (.002)          & \textbf{0.80} (.002) & \textbf{0.80} (.001) \\
              & sof       & 68               & 4               & 0.78 (.007)          & 0.78 (.001)          & \textbf{0.79} (.004) & \textbf{0.79} (.002) \\
              & w. mean   &                  &                 & 0.799                & 0.787                & 0.812                & 0.804                \\
              & w. std    &                  &                 & 0.030                & 0.032                & 0.030                & 0.029                \\
    \midrule
    Hold-Out  & dodh      & 25               & 14              & 0.81 (.012)          & 0.79 (.020)          & 0.83 (.012)          & \textbf{0.84} (.004) \\
    test sets & dodo      & 55               & 10              & \textbf{0.79} (.007) & 0.74 (.016)          & \textbf{0.79} (.008) & 0.78 (.010)          \\
              & isruc-sg1 & 100              & 8               & 0.77 (.002)          & 0.77 (.002)          & \textbf{0.78} (.003) & \textbf{0.78} (.007) \\
              & isruc-sg2 & 16               & 8               & \textbf{0.74} (.002) & 0.73 (.002)          & \textbf{0.74} (.001) & \textbf{0.74} (.003) \\
              & isruc-sg3 & 10               & 8               & \textbf{0.77} (.002) & \textbf{0.77} (.004) & 0.76 (.007)          & \textbf{0.77} (.005) \\
              & mass-c1   & 53               & 21              & 0.72 (.009)          & 0.71 (.004)          & \textbf{0.74} (.010) & 0.73 (.004)          \\
              & mass-c3   & 62               & 23              & 0.78 (.005)          & 0.77 (.004)          & \textbf{0.80} (.010) & \textbf{0.80} (.001) \\
              & svuh      & 25               & 4               & \textbf{0.74} (.004) & 0.73 (.006)          & \textbf{0.74} (.003) & \textbf{0.74} (.005) \\
              & w. mean   &                  &                 & 0.768                & 0.752                & 0.777                & 0.774                \\
              & w. std    &                  &                 & 0.028                & 0.025                & 0.027                & 0.029                \\
  \end{tabular}
  \caption{
    Model performance of \USleep{} and AnySleep with different placements of the channel-attention modules on the in-distribution and hold-out test sets, quantified by macro F1 scores.
    In early fusion, the attention modules were placed at the beginning of the network; in late fusion, they were placed at the end.
    Scores were calculated using all available channels for each dataset ($N_{\text{Ch}}$, see Supplementary Table~\ref{sup-tab:channels} for detailed channel overview) and then weighted by the number of test recordings ($N_{\text{Rec}}$) to obtain weighted means (w. mean) and weighted standard deviations (w. std).
    For the \USleep{} baseline, we followed Perslev~et~al.~\cite{Perslev2021}, generating predictions for all (EOG, EEG) channel pairs and  combining them by majority voting.
    Each architecture was trained three times with different random seeds.
    For each dataset, we report the mean and standard deviation (in parentheses) of macro F1 scores across these three runs, with the best score shown in bold.
  }
  \label{tab:architectures}
\end{table*}

The channel-attention modules that allow AnySleep to handle variable channel combinations can be placed at different network depths (see Section~\ref{ssec:model}).
Network layers before these modules operate on individual channels, whereas layers following them operate on the combined, cross-channel features.
Consequently, placing channel-attention modules at the start of the network architecture biases the model towards cross-channel features, whereas placing them near the end of the model emphasizes channel-wise features.
In the baseline AnySleep configuration, we located the attention modules mid-network to balance these two extremes.
To gain a better understanding of this design choice, we implemented two variants: early fusion and late fusion, in which the intermediate channel-attention modules were replaced by a single module at the beginning or end of the network, respectively.

We compared AnySleep to its early and late fusion variants and to the original \USleep{} architecture.
Across the four architectures, the baseline AnySleep model achieved the highest macro F1 scores on most test datasets (see Table~\ref{tab:architectures}).
On the in-distribution test sets, AnySleep achieved an average macro F1 score of 0.812 (weighted by the number of recordings in each dataset), slightly outperforming late fusion (0.804) and \USleep{} (0.799), while early fusion underperformed on most datasets and attained the lowest average macro F1 score (0.787).
Similar trends were observed on the hold-out test sets:
AnySleep achieved an average macro F1 of 0.777, compared with 0.774 for late fusion, 0.768 for \USleep{}, and 0.752 for early fusion.
All architectures showed modest performance drops from in-distribution to hold-out datasets, providing an empirical estimate of the performance loss to expect when these models are deployed in centers different from those providing the training data.
We also observed modest variability between training runs, suggesting that stochastic factors such as training data sampling and weight initialization could influence model performance and could be further controlled through improved training procedures.

\section{Discussion}

In this work, we presented AnySleep, a deep neural network that accepts any combination of EEG and EOG channels and produces sleep stage predictions at adjustable temporal resolution.
AnySleep was trained on a heterogeneous collection of 13 datasets covering diverse subject populations, clinical centers, and recording setups.
On held-out test data from studies not used in training, AnySleep matched or slightly exceeded the state-of-the-art performance of \USleep{}~\cite{Perslev2021} (see Table~\ref{tab:architectures}) and achieved scores comparable to other recent models validated on independent cohorts~\cite{Guillot2021,Olesen2020,Vallat2021,Hanna2023}; per-dataset and per-stage results as well as confusion matrices are reported in Supplementary Tables~\ref{sup-tab:micros-anysleep}--\ref{sup-tab:micros-usleep} and Supplementary Figure~\ref{sup-fig:confusion-matrices}.
Moreover, sleep metrics derived from AnySleep's predicted hypnograms were comparable to those derived from \USleep{}'s predictions, with modest mean deviations to expert-derived values for most metrics (see Supplementary Table~\ref{sup-tab:hyp-derived-metrics}).
The performance of AnySleep was further validated by its ability to match or exceed typical expert--expert agreement (see Supplementary Figure~\ref{sup-fig:inter-rater}).

A key property of AnySleep is its ability to handle flexible EEG and EOG channel configurations through channel-attention modules.
When evaluated across two-channel permutations, model performance remained stable, and AnySleep maintained strong performance even in single-channel configurations (Figure~\ref{fig:channel_comb_matrix}).
Consistent with prior studies~\cite{Chambon2018,Phan2019,Guillot2021}, performance improved with additional channels, reflecting the benefit of increased spatial resolution.
In contrast, \USleep{} expects a fixed input format of one EEG and one EOG channel~\cite{Perslev2021}, and its performance dropped substantially when the input channels deviated from this configuration (Figure~\ref{fig:channel_comb_matrix}).
A recent extension of \USleep{} partially addressed this limitation by reducing the number of required channels to a single EEG or EOG channel~\cite{Rossi2025b}.
However, when evaluating recordings with more than one EEG and one EOG channel, \USleep{} and its advancements require multiple evaluations of different channels followed by majority voting over the resulting predictions (Figure~\ref{fig:compute-reqs}).
While such post-hoc aggregation can, in principle, exploit information from multiple channels, our results indicate that it does not fully substitute for model components explicitly designed to learn cross-channel relationships.

Motivated by the hypothesis that post-hoc aggregation underutilizes cross-channel relationships, we assessed how fusing channels at different points in the network affects model performance using two variants of AnySleep.
The early fusion variant emphasizes cross-channel features, relying on a small channel-wise feature extractor before combining channels.
This configuration performed substantially worse than the original AnySleep architecture (Table~\ref{tab:architectures}), suggesting that limited channel-wise capacity impaired the extraction of informative per-channel features and hindered channel combination.
The late fusion variant focuses on extracting channel-wise features, combining them only shortly before the final classification layers.
This variant outperformed early fusion but did not reach the performance of the baseline AnySleep model (Table~\ref{tab:architectures}), indicating that optimal performance requires a balance between channel-specific and cross-channel features.
AnySleep achieves this balance by adaptively combining channel information at multiple network depths, which allows the model to shift its focus across channels and modalities at different feature abstraction levels (Figure~\ref{fig:attention-weights}).
Such ``gradual'' fusion strategies~\cite{Stahlschmidt2022} are consistent with the idea that EEG and EOG channels provide complementary information at different temporal and spatial scales.
To our knowledge, this is the first study to investigate these strategies for sleep staging models, which have predominantly relied on early or late fusion schemes to handle variable channel configurations~\cite{Guillot2021,Shi2024,Rossi2025}.

Beyond handling heterogeneous montages, AnySleep predicts sleep stages at temporal resolutions of up to 128~Hz, which allows the model to capture short-lived sleep events such as micro-arousals that are often obscured in conventional 30-s staging.
Consistent with the literature~\cite{Younes2015,BrinkKjaer2020}, we found that expert-annotated micro-arousals, typically described as short and sudden awakenings~\cite{BrinkKjaer2020,Hermans2022,Younes2015}, were rarely represented as Wake in standard 30-s epoch scoring (Figure~\ref{fig:arousals}b).
As we increased the temporal resolution of the predicted stages, the proportion of expert-annotated arousals that aligned with Wake predictions increased, indicating that AnySleep's high-frequency outputs encode these brief events.
A simple rule-based detector applied to these predictions identified up to 52.1\% of expert-annotated arousals, with a maximum IoU F1 score of 0.442, and achieved optimal performance for timescales of 2--8~s, which aligns with typical arousal durations of 3--15~s~\cite{Berry2020,BrinkKjaer2020}.
Repeating this analysis with \USleep{} yielded qualitatively similar results (with predictions identifying up to 50.2\% of expert-annotated arousals), suggesting that sensitivity to arousals is a broader property of high-resolution sleep staging.
AnySleep extends this capability to heterogeneous montages without fixed channel requirements, which is important for multi-center applications.

To further probe the information contained in high-frequency sleep stages, we extracted transition patterns from AnySleep's predicted hypnograms and trained random forest classifiers to discriminate between healthy controls and patients with obstructive sleep apnea (OSA), narcolepsy type 1 (NT1), or insomnia.
Prediction performance improved for all three disorders as the temporal resolution of the sleep stages increased (Figure~\ref{fig:high-freq}), supporting the notion that OSA, NT1, and insomnia influence sleep dynamics on short timescales and that AnySleep can encode these dynamics in its outputs.
This is consistent with prior work reporting that finer temporal scales can reveal increasingly disrupted sleep continuity in OSA patients~\cite{Korkalainen2021}, and with the observation that 5-s resolution sleep-stage features are particularly informative for discriminating patients with REM sleep behavior disorder (RBD) from controls~\cite{Cesari2021}.
Beyond sleep disorders, the prediction of physiological characteristics, such as age and sex, also benefited from higher temporal resolutions (see Supplementary Figure~\ref{sup-fig:high-freq-age-sex}).
Notably, the optimal temporal resolution differed across tasks, with OSA, NT1, and insomnia discriminators performing best at around 0.1, 0.2, and 1.0~s, respectively, whereas arousal detection was best at 2--8~s.
Our feature-selection analysis further indicated that each disorder was characterized by a distinct set of transition triplets, such as N2--Wake disruptions in OSA, REM--wake instability in NT1, and wake intrusions into NREM sleep in insomnia.
While preliminary, these investigations demonstrate how models that represent sleep dynamics across a flexible range of temporal resolutions may serve as an interpretable tool for studying links between sleep disorders and sleep architecture.

Despite the promising findings, several methodological limitations of this study warrant consideration.
First, AnySleep was trained and evaluated for sleep staging using 30-s epochs, and high-frequency predictions were assessed indirectly, through their overlap with annotated arousals and their utility for predicting various sleep-related disorders.
As a result, the absolute accuracy of high-frequency predictions remains uncertain.
Investigations of \USleep{} have demonstrated that performance decreases when evaluated against expert annotations at 5-s resolution, while also noting substantial human inter-rater disagreement, likely due to a lack of standardized scoring rules for shorter sleep stages~\cite{Follin2025b,Follin2025}.
Second, we observed a slight decrease in performance when moving from in-distribution data to held-out test datasets, in line with previous reports on distributional shifts in sleep staging~\cite{Phan2022}.
Given that model performance is constrained by human inter-rater variability~\cite{vanGorp2022,Bechny2025}, further performance gains may be increasingly incremental and are likely to depend on improved annotations, strategies to address distribution shifts~\cite{Chambon2018b,Phan2021}, and broader, more representative training cohorts.
This is particularly relevant for mobile recording devices, which tend to be more prone to artifacts and have shown discrepancies in sleep statistics compared with in-laboratory polysomnography systems~\cite{Esfahani2023,Markov2025}.
Third, we observed modest variability in model performance across training runs, which may suggest that data and channel sampling strategies could be further improved to increase performance, particularly for underrepresented sleep stages such as N1.
Finally, although the AnySleep model is relatively small ($\approx$~12~MB), its computational demands may still be challenging for deployment on low-power consumer hardware where on-device sleep staging is desirable for privacy reasons.

Our findings suggest several directions for future research.
First, developing expert-annotated datasets with sleep stages labeled at high temporal resolution would enable direct evaluation of model performance in the high-frequency regime instead of relying on proxy measures.
Such datasets should ideally include multiple raters and consensus annotations to quantify inter-rater variability at short timescales.
Second, more extensive studies of high-frequency sleep annotations and their applications are warranted.
Potential use cases include the development of biomarkers for sleep disorders such as OSA~\cite{BrinkKjaer2020,Korkalainen2021}, NT1~\cite{Brink2025,Christensen2015}, REM sleep behavior disorder (RBD)~\cite{Cesari2021}, and insomnia~\cite{Moul2007}, as well as for the early detection of neurodegenerative diseases associated with sleep changes~\cite{Cesari2021c}.
Our results provide initial evidence that high-frequency stages may carry relevant information for some of these tasks, but dedicated studies will be needed to validate these findings.
Third, as with most studies based on large, well-annotated polysomnography cohorts, our analysis is retrospective by necessity.
Although evaluations on held-out test datasets provide estimates of expected model performance in practice, prospective validation in routine clinical workflows remains necessary.
Fourth, better methods may exist for aggregating AnySleep's high-frequency representations into 30-s sleep stages, beyond the average pooling currently used in the segment classifier or simple voting schemes (see Supplementary Figure~\ref{sup-fig:high-freq-agg}).
Such aggregations could also provide a more reliable form of uncertainty quantification than interpreting model output probabilities as confidence values~\cite{Brink2025}, thereby improving the basis for hypnodensity analyses~\cite{Stephansen2018,Anderer2023}.
Finally, incorporating training data from mobile recording devices and reducing the model's computational footprint would enable the study of sleep dynamics on short timescales in large cohort studies outside clinical settings.
Given its ability to handle diverse and sparse channel configurations, AnySleep appears well-suited for extension to wearable EEG systems, provided that sufficient high-quality data is collected and domain-specific challenges, such as increased artifact levels, are addressed.

In conclusion, AnySleep provides robust sleep staging across heterogeneous input montages, representing a further step towards large-scale, harmonized sleep staging across studies and centers.
Beyond conventional 30-s scoring, AnySleep generates sleep stage predictions at higher temporal resolutions, which, in our analysis, revealed information obscured by traditional staging.
The interpretation and principled use of such high-resolution predictions remain an emerging area of sleep research.
We expect AnySleep and comparable models to help researchers characterize short-scale sleep dynamics and, in the longer term, to facilitate the development of novel analytical methods and clinically useful biomarkers for sleep and neurological disorders.
To support these developments, we publicly release AnySleep's source code and trained model weights, complemented by our SomnoBot platform, which offers an easy-to-use interface for users without a programming background.

\section{Methods}
\label{sec:methods}

\subsection{Data}
\label{ssec:data}

We trained deep neural networks for automatic sleep staging, a multi-class classification problem where short segments of EEG and EOG data (sleep epochs) are mapped to one of five stages (Wake, N1, N2, N3, REM).
The models were trained and evaluated on data from 28 datasets (20,589 overnight recordings, $\approx$ 200,000~h of EEG/EOG data) from multiple studies and clinics (Table~\ref{tab:datasets}), covering healthy participants and patients with various sleep and medical disorders.
This study is a secondary analysis of previously published, de-identified datasets.
No new data were collected from human participants by the authors.
All datasets were accessed and used in accordance with their applicable data-use agreements and licenses.
Therefore, no additional ethics approval was required for the present secondary analysis.
All datasets included at least one EEG and one EOG channel.
We note that AnySleep could naturally accommodate EMG channels as well, but we refrained from doing so as preliminary experiments indicated no performance improvements, in line with previous findings~\cite{Perslev2021,Hanna2023,Vallat2021}.

\begin{table*}[t]
  \centering
  \begin{tabular}{l|cccrrr}
    Dataset                                          & age in years & sex (F/M)       & BMI         & train rec. & valid rec. & test rec. \\
    \midrule
    abc~\cite{Bakker2018,Zhang2018}                  & 50.0 (9.4)   & 41\% / 59\%     & 37.2 (3.8)  & 97         & 15         & 20        \\
    ccshs~\cite{Rosen2003,Zhang2018}                 & 17.7 (0.4)   & 50\% / 50\%     & 25.1 (5.9)  & 387        & 50         & 78        \\
    cfs~\cite{Redline1995,Zhang2018}                 & 41.4 (19.4)  & 55\% / 45\%     & 32.4 (9.5)  & 569        & 69         & 92        \\
    chat~\cite{Marcus2013,Zhang2018}                 & 7.0 (1.4)    & 51\% / 47\%$^1$ & 19.0 (5.1)  & 1444       & 65         & 128       \\
    dcsm~\cite{Perslev2021c}                         & -            & -               & -           & 190        & 26         & 39        \\
    hpap~\cite{Rosen2012,Zhang2018}                  & 46.5 (11.9)  & 43\% / 57\%     & 37.3 (9.2)  & 178        & 24         & 36        \\
    mesa~\cite{Chen2015,Zhang2018}                   & 69.4 (9.1)   & 54\% / 46\%     & 28.7 (5.6)  & 1904       & 50         & 100       \\
    mros~\cite{Blackwell2011,Zhang2018}              & 77.6 (5.6)   & 0\% / 100\%     & 27.1 (3.8)  & 3714       & 66         & 134       \\
    phys~\cite{Ghassemi2018,Goldberger2000}          & 55.2 (14.3)  & 33\% / 67\%     & -           & 844        & 50         & 100       \\
    sedf-sc~\cite{Kemp2000,Goldberger2000}           & 59.0 (22.0)  & 54\% / 46\%     & -           & 115        & 15         & 23        \\
    sedf-st~\cite{Kemp2000,Goldberger2000}           & 40.2 (17.7)  & 68\% / 32\%     & -           & 30         & 6          & 8         \\
    shhs~\cite{Quan1997,Zhang2018}                   & 62.9 (11.0)  & 53\% / 47\%     & 28.2 (5.1)  & 8227       & 77         & 140       \\
    sof~\cite{Spira2007,Zhang2018}                   & 82.8 (3.1)   & 100\% / 0\%     & 27.7 (4.7)  & 339        & 46         & 68        \\
    \midrule
    dodh~\cite{Guillot2020}                          & 35.3 (7.5)   & 24\% / 76\%     & 23.8 (3.4)  & -          & -          & 25        \\
    dodo~\cite{Guillot2020}                          & 45.6 (16.5)  & 36\% / 64\%     & 29.6 (6.4)  & -          & -          & 55        \\
    isruc-sg1~\cite{Khalighi2016}                    & 51.1 (15.9)  & 44\% / 56\%     & -           & -          & -          & 100       \\
    isruc-sg2~\cite{Khalighi2016}                    & 46.9 (17.5)  & 25\% / 75\%     & -           & -          & -          & 16        \\
    isruc-sg3~\cite{Khalighi2016}                    & 39.6 (9.6)   & 10\% / 90\%     & -           & -          & -          & 10        \\
    mass-c1~\cite{OReilly2014}                       & 63.6 (5.3)   & 36\% / 64\%     & -           & -          & -          & 53        \\
    mass-c3~\cite{OReilly2014}                       & 42.5 (18.9)  & 55\% / 45\%     & -           & -          & -          & 62        \\
    svuh~\cite{McNicholas2004,Goldberger2000}        & 50.0 (9.4)   & 16\% / 84\%     & 31.6 (3.9)  & -          & -          & 25        \\
    \midrule
    cnc~\cite{Andlauer2012,Stephansen2018}           & 28.5 (16.9)  & 49\% / 51\%     & 23.2 (11.5) & -          & -          & 165$^2$   \\
    dhc~\cite{Christensen2017,Stephansen2018}        & 33.4 (14.8)  & 50\% / 50\%     & 24.8 (4.9)  & -          & -          & 41$^2$    \\
    fhc~\cite{Stephansen2018}                        & 28.8 (15.2)  & 41\% / 59\%     & 24.4 (8.1)  & -          & -          & 53$^2$    \\
    ihc~\cite{Pizza2015,Stephansen2018}              & 33.7 (17.6)  & 43\% / 57\%     & -           & -          & -          & 70$^2$    \\
    khc~\cite{Andlauer2013,Hong2006,Stephansen2018}  & 29.1 (13.2)  & 41\% / 59\%     & 24.1 (4.3)  & -          & -          & 80$^2$    \\
    ssc~\cite{Andlauer2013,Moore2014,Stephansen2018} & 45.4 (13.8)  & 41\% / 59\%     & 23.9 (6.5)  & -          & -          & 246$^2$   \\
    cap~\cite{Terzano2001,Goldberger2000}            & 42.5 (15.6)  & 56\% / 44\%     & -           & -          & -          & 25$^3$    \\
  \end{tabular}
  \caption{
    Datasets used in this study, with demographics and the number of recordings per training, validation, and test split.
    The table is grouped into in-distribution datasets used for training, validation, and testing (top); hold-out datasets used only for testing (middle); and datasets used only to evaluate high-frequency sleep staging (bottom; see Section~\ref{ssec:high-freq-results}).
    Age and body mass index (BMI) are reported as mean (standard deviation).
    $^1$Sex was unspecified for the remaining 2\% of recordings.
    $^2$Count restricted to healthy controls and narcolepsy type 1 (NT1) patients; demographics taken from Stephansen~et~al.~\cite{Stephansen2018}, also include patients diagnosed with other hypersomnias.
    $^3$Restricted to healthy controls and insomnia patients.
  }
  \label{tab:datasets}
\end{table*}

The datasets were grouped into 13 in-distribution datasets and 15 hold-out datasets.
Only in-distribution datasets contributed to model training and validation, while hold-out datasets were reserved exclusively for testing the traditional 30-s sleep staging (8 datasets) and high-frequency capabilities (7 datasets).
Within each in-distribution dataset, subjects were partitioned into training, validation, and test subsets, with 10\% of subjects (up to 50 per dataset) used for validation, 15\% (up to 100 per dataset) for testing, and the remainder for training.
This yielded 18,038 training recordings, 559 validation recordings, and 1992 test recordings (966 from the in-distribution group, 346 from the 30-s hold-out group, and 680 from the high-frequency hold-out group; see Table~\ref{tab:datasets}).
To ensure comparability with the results reported by Perslev et al.~\cite{Perslev2021}, our data split closely followed their protocol~\cite{Perslev2021b}, except for the high-frequency hold-out group, which was not included in their study.
In addition to the 30-s and high-frequency hold-out groups, we evaluated a high-density EEG dataset (ANPHY~\cite{Wei2024}) to assess AnySleep's performance in recordings with substantially larger channel counts (Supplementary Table~\ref{sup-tab:high-density}).
Across all datasets, we excluded 32 recordings flagged as problematic by the data providers or lacking EEG/EOG channels (see code repository available at \url{https://github.com/dslaborg/AnySleep}).

All recordings used for training or to evaluate 30-s staging were scored by expert annotators into 30-s sleep epochs according to either the AASM~\cite{Berry2020} or the Rechtschaffen and Kales~\cite{Rechtschaffen1968} rules.
To harmonize labels, we remapped stage N4 to N3.
We did not remove epochs labeled outside the five main stages (e.g., movement, artifacts) to prevent discontinuities between non-consecutive epochs and to familiarize the model with artifacts, but excluded such epochs from loss computation and evaluation metrics (Section~\ref{ssec:training}).
For the DODO and DODH datasets, which each provide annotations of five independent scorers per recording, we derived consensus labels by majority voting.
Scorers were ranked per recording by their mean agreement with the other scorers, and only the four most reliable scorers contributed to the vote, with ties resolved in favor of the highest-ranked scorer~\cite{Guillot2020}.
Some datasets additionally contain event annotations and subject-level metadata, which we used in downstream analyses, namely arousal annotations in MASS C1 and C3, and clinical indicators for patient versus healthy control status, in DODO, DODH, CNC, DHC, FHC, IHC, KHC, SSC, and CAP\@.
For all datasets other than DODO and DODH, annotations were available from a single expert scorer per recording.

For preprocessing, all EEG and EOG signals were resampled to 128~Hz using polyphase filtering.
Then, we normalized the amplitudes of each recording and channel by subtracting the median and dividing by the interquartile range of the amplitude distribution.
To minimize outliers, the normalized signal was clipped to the range $[-20,20]$.
Following \USleep{}~\cite{Perslev2021}, we did not apply bandpass filtering, as our preliminary experiments with the \USleep{} architecture on bandpass-filtered data did not yield notable performance improvements.

\subsection{Model}
\label{ssec:model}

\begin{figure*}[t]
  \centering
  \includegraphics[width=\linewidth]{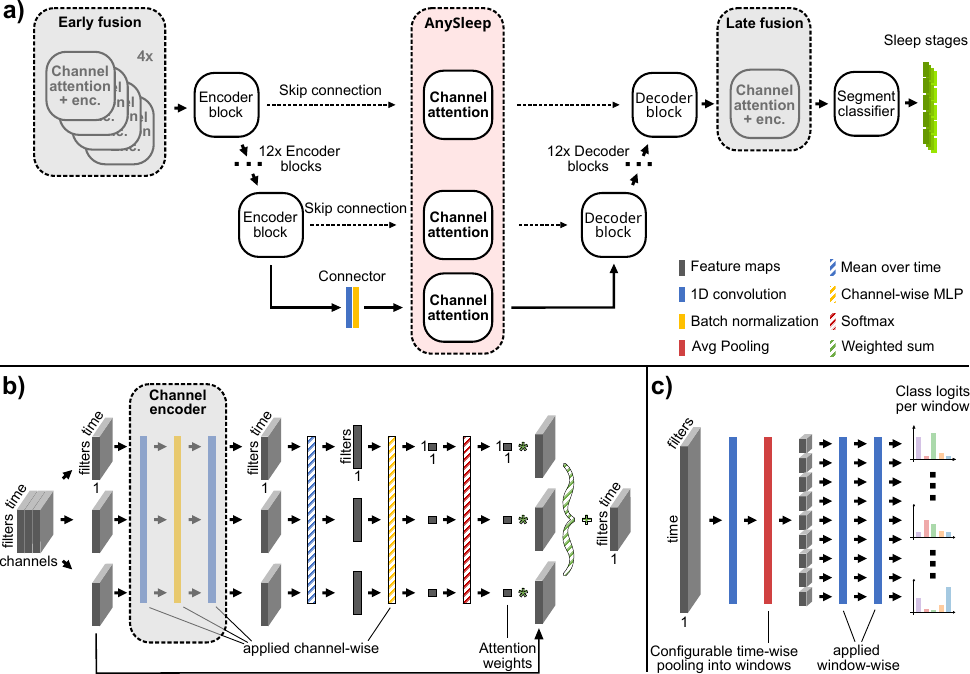}
  \caption{
    Model architecture of AnySleep.
    \textbf{(a)} gives an overview of AnySleep with its two variants: early fusion and late fusion (see Section~\ref{ssec:fusion_placement}).
    In the two variants, the attention modules of AnySleep (highlighted in red) were extended with a channel encoder and moved before the first encoder block or after the last decoder block, respectively (greyed out modules).
    \textbf{(b)} Architecture of the channel-attention modules.
    The channel encoder serves as an additional channel-wise feature extractor in the early and late fusion variants.
    \textbf{(c)} Architecture of the segment classifier with the configurable average pooling layer that allows for the prediction of high-frequency sleep stages at flexible resolution.
  }
  \label{fig:architecture}
\end{figure*}

The AnySleep architecture is a U-Net-style encoder-decoder network for multi-channel EEG/EOG segmentation.
We adopted the backbone configuration from \USleep{}~\cite{Perslev2021}, with 12 encoder blocks, a connector, and 12 decoder blocks connected by skip connections (Figure~\ref{fig:architecture}a).
The output of the last decoder block is passed to a convolutional \emph{segment classifier} (Figure~\ref{fig:architecture}c) with a temporal average-pooling layer.
The kernel size and stride of this pooling layer determine the effective temporal resolution of the output.
During training, we set this pooling window to 30~s (3840 samples at 128~Hz) to match the expert 30-s annotations.
At inference, we varied the pooling kernel and stride to obtain higher-resolution sleep stage predictions from the same backbone.
We additionally evaluated soft and hard majority voting as alternatives to average pooling for aggregating these high-frequency predictions into 30-s labels (see Supplementary Figure~\ref{sup-fig:high-freq-agg}).

To support flexible input channel configurations, we introduced channel-attention modules that combined information across any number of channels (Figure~\ref{fig:architecture}b).
Layers before a channel-attention module operate on each channel separately, and layers after the module operate on a fused multi-channel representation.
In the baseline AnySleep model, we inserted 13 channel-attention modules at different depths between the encoder and decoder blocks (Figure~\ref{fig:architecture}a).
For an input with $C$ channels, each attention module receives a feature map $m\in \mathbb{R}^{C\times F\times T}$, where $F$ denotes the number of convolutional filters and $T$ the temporal dimension.
Inspired by the attention mechanism described by Guillot~et~al.~\cite{Guillot2021}, our attention modules split these maps into channel-wise feature maps $m_i\in \mathbb{R}^{F\times T}$, which are averaged over the time dimension $T$, yielding $\overline{m}_i = \frac{1}{T}\sum_{t=1}^{T} m_{i,t},\: \overline{m}_i\in \mathbb{R}^{F}$.
The $\overline{m}_i$ are then passed through a multi-layer perceptron (MLP) with one hidden layer (40 units), batch normalization, a ReLU activation, and a single output unit.
The resulting scalars are normalized across channels with a softmax layer to obtain normalized attention weights $w_i\in \mathbb{R}$ with $\sum_{i=1}^{C} w_i = 1$.
These attention weights are used to calculate the weighted sum over the channel-wise feature maps $m_i$, yielding an aggregated feature map $m_{\text{agg}}=\sum_{i=1}^{C} w_i m_i,\: m_{\text{agg}}\in \mathbb{R}^{F\times T}$ without the channel dimension.

In the early and late fusion variants of AnySleep (Section~\ref{ssec:fusion_placement}), the attention modules were replaced with a single attention module positioned either before the first encoder block or after the last decoder block, respectively (Figure~\ref{fig:architecture}a,b).
Additionally, these modules were extended with a channel encoder that serves as an additional channel-wise feature extractor.
The feature extractor consists of two convolutional layers (32 filters, kernel sizes 64 and 9, strides 32 and 1, respectively) with an ELU activation and batch normalization between them.
For the early fusion architecture, we implemented a \emph{multi-head attention} mechanism~\cite{Vaswani2017,Guillot2021} with four parallel channel-attention modules, yielding $C_{\text{virtual}}=4$ fused feature maps (``virtual channels''), $m_{\text{fused}}\in \mathbb{R}^{C_{\text{virtual}}\times F\times T}$.
This increases the number of input channels of the first encoder block from 1 to 4.
In contrast, the baseline AnySleep model and the late fusion variant use single-head attention, as preliminary experiments did not show consistent benefits of multi-head fusion in this setting.

\begin{figure}[h]
  \centering
  \includegraphics[width=\linewidth]{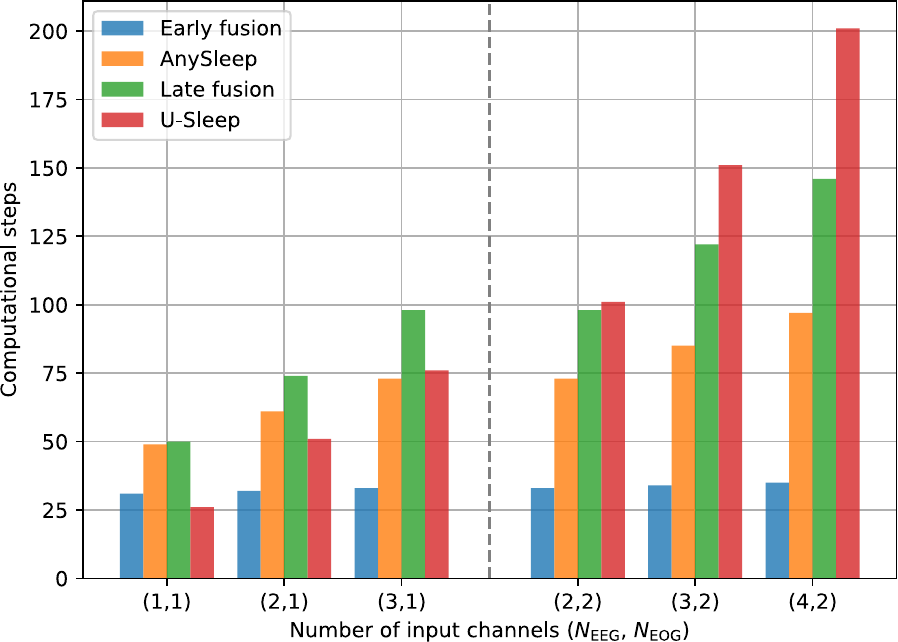}
  \caption{
    Number of computational steps required for AnySleep, its two variants, and \USleep{} as a function of the number of EEG and EOG channels.
    We defined the number of computational steps as the count of individual model components (channel encoders, attention modules, encoder blocks, decoder blocks, and segment classifier) applied to the input data.
  }
  \label{fig:compute-reqs}
\end{figure}

The introduction of channel-attention modules reduced the computational requirements needed to evaluate recordings with increasing channel numbers compared to \USleep{}.
We approximated the compute required to score a single recording with $N_{\text{EEG}}$ EEG channels and $N_{\text{EOG}}$ EOG channels by counting how often each network component (encoder block, decoder block, channel encoder, attention module, segment classifier) needed to be evaluated (Figure~\ref{fig:compute-reqs}).
In AnySleep and its two variants, components before the attention modules are applied independently to each input channel, so the number of component evaluations increases linearly with the channel count.
The steepness of this increase depends on the number of channel-specific components (largest for late fusion, smallest for early fusion).
In contrast, \USleep{} handles increasing channel numbers by performing additional full model evaluations with subsequent majority voting.
In the original version of \USleep{}~\cite{Perslev2021}, a recording with $N_{\text{EEG}}$ EEG and $N_{\text{EOG}}$ EOG channels requires separate evaluations for all (EOG, EEG) pairs (i.e., \mbox{$N_{\text{EEG}} \cdot N_{\text{EOG}}$} evaluations), leading to a quadratic increase in computational cost with the number of channels.
This scaling makes \USleep{} less compute-efficient relative to AnySleep in datasets with large channel numbers, especially when more than one EOG channel is available.
In contrast, introducing additional channel-attention components only modestly increased the models' parameters, with AnySleep, early fusion, and late fusion containing 3,157,856 (+1.4\% compared to \USleep{}'s 3,114,337 parameters), 3,131,601 (+0.6\%), and 3,137,356 (+0.7\%) parameters, respectively.

\subsection{Training}
\label{ssec:training}

Each training sample consisted of a sequence of 35 contiguous 30-s sleep epochs (17.5~min) to leverage the models' receptive fields spanning 14.36~min.
For each training sample, the models predicted sleep stages for every epoch in the sequence, and we minimized the average cross-entropy loss across these epochs.
Epochs annotated as artifacts or unknowns by the experts (see Section~\ref{ssec:data}) were excluded from the loss calculation.
Optimization used the AMSGrad variant of Adam~\cite{Reddi2018} with a fixed learning rate of $10^{-5}$, following prior work~\cite{Fiorillo2023,Rossi2025b}, and a batch size of 64 (reduced to 32 for the late-fusion AnySleep variant due to memory constraints).
We trained for a maximum of 10,000 training epochs and applied early stopping if the macro F1 score on the validation data did not improve for 100 consecutive training epochs.
Each training epoch consisted of 443 gradient updates, which corresponded to approximately $10^6$ sleep epochs for a batch size of 64 and a sequence length of 35.

We generated training samples by stratified sampling over sleep stages and datasets, similar to Perslev~et~al.~\cite{Perslev2021}.
For each training sample, we first uniformly sampled a sleep stage $c \in \{\text{Wake, N1, N2, N3, REM}\}$.
We then sampled a dataset $d$ from the $N_d$ available training datasets with probability
\begin{equation}
  p_d = \alpha \frac{1}{N_d} + (1-\alpha) \frac{N_{\text{rec}_d}}{\sum_{i=1}^{N_d} N_{\text{rec}_i}},
\end{equation}
where $N_{rec_d}$ is the number of recordings in dataset $d$ and $\alpha$ is a hyperparameter controlling the balance between equal weighting of datasets and weighting by dataset size.
We set $\alpha = 0.5$ to ensure that recordings from smaller datasets are neither under- nor overrepresented.
From the selected dataset, we uniformly sampled a recording and then uniformly sampled a sleep epoch of class $c$ within that recording.
If no such epoch was present, the procedure was repeated from the dataset selection step.
Once a sleep epoch of class $c$ was selected, it was placed at a random position within a 35-epoch sequence by uniformly sampling the number of preceding epochs from $\{0, \ldots, 34\}$.

To expose the model to a wide range of channel numbers and combinations during training, we used stochastic channel subsampling inspired by Guillot~et~al.~\cite{Guillot2021}.
For each batch, we first randomly sampled a number $n$ of $1, \dots, N_{\text{ch}}$ channels with probability
\begin{equation}
  p_n = \left(\sum_{i=1}^{N_{\text{ch}}} \frac{n}{i}\right)^{-1},
\end{equation}
where $N_{\text{ch}}=10$ is the maximum number of channels across all train recordings (see in-distribution datasets in Table~\ref{tab:architectures}).
For each recording in the batch, we then uniformly selected $n$ of its available channels, independent of the channel type (EEG or EOG).
If a recording had less than $n$ channels, sampling was performed with replacement; otherwise, channels were sampled without replacement.

In contrast to the \USleep{} training pipeline~\cite{Perslev2021}, we did not apply data augmentation.
Preliminary experiments indicated that the masking-based augmentations used in \USleep{} did not improve AnySleep's performance and slightly degraded its high-frequency sleep staging performance.
To obtain a directly comparable baseline, we retrained \USleep{} using the same sampling strategy, optimization hyperparameters, and stopping criteria as for AnySleep but replaced the channel subsampling scheme with random selection of an (EOG, EEG) pair for each training sample.
Our retrained \USleep{} models closely reproduced the originally reported performance, with weighted mean MF1 scores across all test data of 0.79 (see Table~2 in Perslev~et~al.~\cite{Perslev2021}) and 0.791 for the original and retrained models, respectively, well within the variability observed across training runs.

\subsection{Evaluation}
\label{ssec:evaluation}

We assessed sleep staging performance by calculating macro F1 (MF1) scores, which are defined as the unweighted average of the per-stage F1 scores across the five sleep stages~\cite{Tharwat2021},
\begin{equation}
  \overline{F_1} = \frac{1}{5} \sum_{i=1}^{5} \frac{2 \cdot TP_i}{2 \cdot TP_i + FP_i + FN_i},
\end{equation}
where $TP_i$, $FP_i$, and $FN_i$ are the number of true positives, false positives, and false negatives of sleep stage $i$, respectively.
Only sleep epochs annotated as Wake, N1, N2, N3, or REM were included in these calculations, while epochs annotated as artifacts or unknowns were excluded (see Section~\ref{ssec:data}).
Macro F1 scores were calculated either recording-wise or dataset-wise by aggregating $TP_i$, $FP_i$, and $FN_i$ over all epochs of a recording (Section~\ref{ssec:channel-robustness}) or dataset (Section~\ref{ssec:fusion_placement}), respectively.

Because both AnySleep and \USleep{} accept input signals of variable length, we passed full recordings to the models, ensuring that each sleep epoch prediction could exploit the models' full temporal receptive field without resorting to segmenting the input into overlapping windows.
AnySleep processed all available channels jointly in a single forward pass.
For \USleep{}, following Perslev~et~al.~\cite{Perslev2021}, we evaluated the model on every available (EOG, EEG) channel pair and combined the resulting per-epoch predictions by majority voting, breaking ties at random.
Unless specified otherwise, all evaluations used all available EEG and EOG channels of a recording.

\subsection{High-frequency sleep stages}
\label{ssec:high-freq}

AnySleep outputs sleep stages at temporal resolutions of up to 128~Hz (3840 predictions per 30-s epoch; see Section~\ref{ssec:model}).
For the high-frequency analyses, we evaluated 14 temporal resolutions ranging from 1 to 3840 predictions per epoch (1, 2, 4, 8, 16, 32, 64, 128, 256, 384, 640, 960, 1920, and 3840), corresponding to time steps from 30~s down to 0.008~s.

\paragraph{Arousal prediction.}
We used AnySleep's high-frequency predictions to derive candidate arousals for the MASS C1 and C3 datasets (see Section~\ref{ssec:high-freq-results}).
Specifically, we first identified segments of continuous Wake predictions.
We merged segments that were separated by less than 10\% of their merged length, provided the resulting event did not exceed 15~s.
Then, in line with AASM scoring rules~\cite{Berry2020}, we removed segments that contained Wake predictions in the preceding 10~s.
Finally, we defined candidate arousals as segments with a length of 3--15~s, consistent with typical arousal durations~\cite{Berry2020,BrinkKjaer2020}.

To compare predicted and expert-annotated arousals, we calculated Intersection over Union (IoU) precision, IoU recall, and IoU F1 scores.
True positives were defined as pairs of predicted and expert-annotated arousals that overlapped by at least 20\% of their combined length.
Predicted arousals without such a match were counted as false positives, and expert events without matching predictions were counted as false negatives (see Section~\ref{ssec:evaluation}).

\paragraph{Analysis of subject characteristics.}
We used AnySleep's high-frequency predictions to quantify associations between fine-grained sleep dynamics and various disorders (obstructive sleep apnea, narcolepsy type 1, and insomnia).
Following Perslev~et~al.~\cite{Perslev2021}, these analyses were based on ``triplets'' of sleep transitions.
To calculate these features, we first predicted sleep stages for the DODO, DODH, CNC, DHC, FHC, IHC, KHC, SSC, and CAP datasets at varying temporal resolutions with all available EEG and EOG channels (except for DODO and DODH, which were limited to channels shared by the two datasets).
We then restricted the predictions to the interval between the first and last non-Wake 30-s epoch, partitioned this interval into non-overlapping 1.5-h blocks, and, for each block and timescale, counted the absolute number of sleep-stage triplets $(s_i, s_{i+1}, s_{i+2})$ where $s_i \neq s_{i+1}$ and $s_{i+1} \neq s_{i+2}$.
This yielded one feature vector with 80 features per 1.5-h block.

Based on these features, we trained random forest classifiers in a leave-one-subject-out cross-validation setting to distinguish healthy controls from patients with obstructive sleep apnea (OSA), narcolepsy type 1 (NT1), or insomnia.
Each of the three tasks was treated as a binary classification problem.
The random forest models were implemented using the \codeword{RandomForestClassifier} from scikit-learn~\cite{Pedregosa2011}, with \codeword{criterion} set to \codeword{gini} and \codeword{class\_weight} set to \codeword{balanced}.
For each disorder and timescale, we trained 50 of these forests with hyperparameters uniformly sampled from the ranges \codeword{max\_tree\_depth}$\in [2,7]$, \codeword{min\_samples\_leaf}$\in [2,7]$, \codeword{min\_samples\_split}$\in [2,7]$, \codeword{max\_features}$\in \{\text{sqrt}, \text{log2}\}$.
Performance was quantified using macro F1 scores.
Specifically, we aggregated block-level predictions for each subject by majority voting across the subject's 1.5-h blocks, and then computed macro F1 scores based on the resulting subject-level predictions.

To investigate the impact of different triplet features on the performance of the random forest classifiers, we carried out a feature selection analysis for the best-performing timescales (256 predictions per epoch for OSA, 128 for NT1, and 32 for insomnia).
For each of the 150 training runs of a disorder (three AnySleep checkpoints with 50 random forest models each), we started with an empty feature set and iteratively added the triplet that maximally improved validation performance up to a maximum of five features.
We then ranked the triplets by how often they were selected across the 150 runs and treated the highest-ranking triplets as the most influential features.

  { \small
    \section*{Data Availability}
    All datasets analyzed during the current study are publicly available.
    In the following, we list the datasets and the URLs to access them: ABC (\url{https://doi.org/10.25822/nx52-bc11}), CCSHS (\url{https://doi.org/10.25822/cg2n-4y91}), CFS (\url{https://doi.org/10.25822/jmyx-mz90}), CHAT (\url{https://doi.org/10.25822/d68d-8g03}), DCSM (\url{https://doi.org/10.17894/ucph.282d3c1e-9b98-4c1e-886e-704afdfa9179}), HPAP (\url{https://doi.org/10.25822/xmwv-yz91}), MESA (\url{https://doi.org/10.25822/n7hq-c406}), MrOS (\url{https://doi.org/10.25822/kc27-0425}), Phys (\url{https://doi.org/10.13026/6phb-r450}), SEDF-ST and SEDF-SC (\url{https://doi.org/10.13026/C2X676}), SHHS (\url{https://doi.org/10.25822/ghy8-ks59}), SOF (\url{https://doi.org/10.25822/e1cf-rx65}), DODO and DODH (\url{https://doi.org/10.5281/zenodo.15900394}), ISRUC SG 1--3 (\url{https://sleeptight.isr.uc.pt/}), MASS C1 and C3 (\url{https://doi.org/10.5683/SP3/OVISPE} and \url{https://doi.org/10.5683/SP3/9MYUCS}), SVUH (\url{https://doi.org/10.13026/C26C7D}), CNC, DHC, FHC, IHC, KHC, and SSC (\url{https://stanfordmedicine.app.box.com/s/r9e92ygq0erf7hn5re6j51aaggf50jly}), CAP (\url{https://doi.org/10.13026/C2VC79}), ANPHY (used in the Supplementary Materials; \url{https://doi.org/10.17605/OSF.IO/R26FH}).
    Information about excluded recordings and the used data split is provided in our GitHub repository: \url{https://github.com/dslaborg/AnySleep}.

    \section*{Code Availability}
    The underlying code, trained model files, and training, validation, and test data splits for this study are available on GitHub and can be accessed via this link: \url{https://github.com/dslaborg/AnySleep}.
    Further instructions on how to reproduce our main experiments and evaluate our trained models on custom datasets are also provided there.
    The software is based on PyTorch (version 2.5.1, \url{https://pytorch.org/}).
    All models were trained on an NVIDIA DGX A100 workstation equipped with eight NVIDIA A100 GPUs.

    \section*{Acknowledgements}
    We are grateful to M. Reißel and V. Sander for providing us with computing resources.
    The Apnea, Bariatric surgery, and CPAP study (ABC Study) was supported by National Institutes of Health grants R01HL106410 and K24HL127307. Philips Respironics donated the CPAP machines and supplies used in the perioperative period for patients undergoing bariatric surgery. The National Sleep Research Resource was supported by the National Heart, Lung, and Blood Institute (R24 HL114473, 75N92019R002).
    The Cleveland Children's Sleep and Health Study (CCSHS) was supported by grants from the National Institutes of Health (RO1HL60957, K23 HL04426, RO1 NR02707, M01 Rrmpd0380-39).
    The Cleveland Family Study (CFS) was supported by grants from the National Institutes of Health (HL46380, M01 RR00080-39, T32-HL07567, RO1-46380).
    The Childhood Adenotonsillectomy Trial (CHAT) was supported by the National Institutes of Health (HL083075, HL083129, UL1-RR-024134, UL1 RR024989).
    The Home Positive Airway Pressure study (HomePAP) was supported by the American Sleep Medicine Foundation 38-PM-07 Grant: Portable Monitoring for the Diagnosis and Management of OSA\@.
    The Multi-Ethnic Study of Atherosclerosis (MESA) Sleep Ancillary study was funded by NIH-NHLBI Association of Sleep Disorders with Cardiovascular Health Across Ethnic Groups (RO1 HL098433). MESA is supported by NHLBI funded contracts HHSN268201500003I, N01-HC-95159, N01-HC-95160, N01-HC-95161, N01-HC-95162, N01-HC-95163, N01-HC-95164, N01-HC-95165, N01-HC-95166, N01-HC-95167, N01-HC-95168 and N01-HC-95169 from the National Heart, Lung, and Blood Institute, and by cooperative agreements UL1-TR-000040, UL1-TR-001079, and UL1-TR-001420 funded by NCATS\@.
    The National Heart, Lung, and Blood Institute provided funding for the ancillary MrOS Sleep Study, ``Outcomes of Sleep Disorders in Older Men,'' under the following grant numbers: R01 HL071194, R01 HL070848, R01 HL070847, R01 HL070842, R01 HL070841, R01 HL070837, R01 HL070838, and R01 HL070839.
    The Sleep Heart Health Study (SHHS) was supported by National Heart, Lung, and Blood Institute cooperative agreements U01HL53916 (University of California, Davis), U01HL53931 (New York University), U01HL53934 (University of Minnesota), U01HL53937 and U01HL64360 (Johns Hopkins University), U01HL53938 (University of Arizona), U01HL53940 (University of Washington), U01HL53941 (Boston University), and U01HL63463 (Case Western Reserve University).
    The Study of Osteoporotic Fractures (SOF) was supported by National Institutes of Health grants (AG021918, AG026720, AG05394, AG05407, AG08415, AR35582, AR35583, AR35584, RO1 AG005407, R01 AG027576-22, 2 R01 AG005394-22A1, 2 RO1 AG027574-22A1, HL40489, T32 AG000212-14).
    Open Access funding enabled and organized by Projekt DEAL.
    The funders played no role in study design, analysis and interpretation of data, or the writing of this manuscript.

    \section*{Author Contributions}
    N.G.\ and S.B.\ conceived the experiments; N.G.\ and J.R.\ conducted the experiments; N.G., J.R., S.M., and S.B.\ analyzed and discussed the
    results; N.G.\ and S.B.\ wrote the first draft of the manuscript; N.G., J.R., S.M., and S.B.\ reviewed the manuscript.
    All authors read and approved the final manuscript.

    \section*{Competing Interests}
    The authors declare no competing interests.
  }

\bibliographystyle{naturemag-doi}

\begin{thebibliography}{10}
\urlstyle{rm}
\expandafter\ifx\csname url\endcsname\relax
  \def\url#1{\texttt{#1}}\fi
\expandafter\ifx\csname urlprefix\endcsname\relax\def\urlprefix{URL }\fi
\expandafter\ifx\csname doiprefix\endcsname\relax\def\doiprefix{DOI: }\fi
\providecommand{\bibinfo}[2]{#2}
\providecommand{\eprint}[2][]{\url{#2}}

\bibitem{Cesari2021b}
\bibinfo{author}{Cesari, M.} \emph{et~al.}
\newblock \bibinfo{journal}{\bibinfo{title}{Interrater sleep stage scoring reliability between manual scoring from two European sleep centers and automatic scoring performed by the artificial intelligence--based Stanford-{STAGES} algorithm}}.
\newblock {\emph{\JournalTitle{J. Clin. Sleep Med.}}} \textbf{\bibinfo{volume}{17}}, \bibinfo{pages}{1237--1247}, \doiprefix\url{https://doi.org/10.5664/jcsm.9174} (\bibinfo{year}{2021}).

\bibitem{Lee2022}
\bibinfo{author}{Lee, Y.~J.}, \bibinfo{author}{Lee, J.~Y.}, \bibinfo{author}{Cho, J.~H.} \& \bibinfo{author}{Choi, J.~H.}
\newblock \bibinfo{journal}{\bibinfo{title}{Interrater reliability of sleep stage scoring: {a} meta-analysis}}.
\newblock {\emph{\JournalTitle{J. Clin. Sleep Med.}}} \textbf{\bibinfo{volume}{18}}, \bibinfo{pages}{193--202}, \doiprefix\url{https://doi.org/10.5664/jcsm.9538} (\bibinfo{year}{2022}).

\bibitem{Nikkonen2023}
\bibinfo{author}{Nikkonen, S.} \emph{et~al.}
\newblock \bibinfo{journal}{\bibinfo{title}{Multicentre sleep-stage scoring agreement in the Sleep Revolution Project}}.
\newblock {\emph{\JournalTitle{J. Sleep Res.}}} \textbf{\bibinfo{volume}{33}}, \doiprefix\url{https://doi.org/10.1111/jsr.13956} (\bibinfo{year}{2023}).

\bibitem{Loomis1936}
\bibinfo{author}{Loomis, A.~L.}, \bibinfo{author}{Harvey, E.~N.} \& \bibinfo{author}{Hobart, G.}
\newblock \bibinfo{journal}{\bibinfo{title}{Electrical potentials of the human brain.}}
\newblock {\emph{\JournalTitle{J. Exp. Psychol.}}} \textbf{\bibinfo{volume}{19}}, \bibinfo{pages}{249--279}, \doiprefix\url{https://doi.org/10.1037/h0062089} (\bibinfo{year}{1936}).

\bibitem{Decat2022}
\bibinfo{author}{Decat, N.} \emph{et~al.}
\newblock \bibinfo{journal}{\bibinfo{title}{Beyond traditional sleep scoring: {m}assive feature extraction and data-driven clustering of sleep time series}}.
\newblock {\emph{\JournalTitle{Sleep Med.}}} \textbf{\bibinfo{volume}{98}}, \bibinfo{pages}{39--52}, \doiprefix\url{https://doi.org/10.1016/j.sleep.2022.06.013} (\bibinfo{year}{2022}).

\bibitem{Lambert2023}
\bibinfo{author}{Lambert, I.} \& \bibinfo{author}{Peter-Derex, L.}
\newblock \bibinfo{journal}{\bibinfo{title}{Spotlight on sleep stage classification based on {EEG}}}.
\newblock {\emph{\JournalTitle{Nat. Sci. Sleep}}} \textbf{\bibinfo{volume}{15}}, \bibinfo{pages}{479--490}, \doiprefix\url{https://doi.org/10.2147/nss.s401270} (\bibinfo{year}{2023}).

\bibitem{Korkalainen2021}
\bibinfo{author}{Korkalainen, H.} \emph{et~al.}
\newblock \bibinfo{journal}{\bibinfo{title}{Detailed assessment of sleep architecture with deep learning and shorter epoch-to-epoch duration reveals sleep fragmentation of patients with obstructive sleep apnea}}.
\newblock {\emph{\JournalTitle{{IEEE J. Biomed. Health Informatics}}}} \textbf{\bibinfo{volume}{25}}, \bibinfo{pages}{2567--2574}, \doiprefix\url{https://doi.org/10.1109/JBHI.2020.3043507} (\bibinfo{year}{2021}).

\bibitem{BrinkKjaer2020}
\bibinfo{author}{Brink-Kjaer, A.} \emph{et~al.}
\newblock \bibinfo{journal}{\bibinfo{title}{Automatic detection of cortical arousals in sleep and their contribution to daytime sleepiness}}.
\newblock {\emph{\JournalTitle{Clin. Neurophysiol.}}} \textbf{\bibinfo{volume}{131}}, \bibinfo{pages}{1187--1203}, \doiprefix\url{https://doi.org/10.1016/j.clinph.2020.02.027} (\bibinfo{year}{2020}).

\bibitem{Malafeev2021}
\bibinfo{author}{Malafeev, A.} \emph{et~al.}
\newblock \bibinfo{journal}{\bibinfo{title}{Automatic detection of microsleep episodes with deep learning}}.
\newblock {\emph{\JournalTitle{Front. Neurosci.}}} \textbf{\bibinfo{volume}{15}}, \doiprefix\url{https://doi.org/10.3389/fnins.2021.564098} (\bibinfo{year}{2021}).

\bibitem{Cesari2021}
\bibinfo{author}{Cesari, M.} \emph{et~al.}
\newblock \bibinfo{journal}{\bibinfo{title}{A data-driven system to identify REM sleep behavior disorder and to predict its progression from the prodromal stage in Parkinson's disease}}.
\newblock {\emph{\JournalTitle{Sleep Med.}}} \textbf{\bibinfo{volume}{77}}, \bibinfo{pages}{238--248}, \doiprefix\url{https://doi.org/10.1016/j.sleep.2020.04.010} (\bibinfo{year}{2021}).

\bibitem{Younes2015}
\bibinfo{author}{Younes, M.} \emph{et~al.}
\newblock \bibinfo{journal}{\bibinfo{title}{Odds ratio product of sleep {EEG} as a continuous measure of sleep state}}.
\newblock {\emph{\JournalTitle{Sleep}}} \textbf{\bibinfo{volume}{38}}, \bibinfo{pages}{641--654}, \doiprefix\url{https://doi.org/10.5665/sleep.4588} (\bibinfo{year}{2015}).

\bibitem{Moul2007}
\bibinfo{author}{Moul, D.~E.} \emph{et~al.}
\newblock \bibinfo{journal}{\bibinfo{title}{Examining initial sleep onset in primary insomnia: {a} case-control study using 4-second epochs}}.
\newblock {\emph{\JournalTitle{J. Clin. Sleep Med.}}} \textbf{\bibinfo{volume}{03}}, \bibinfo{pages}{479--488}, \doiprefix\url{https://doi.org/10.5664/jcsm.26912} (\bibinfo{year}{2007}).

\bibitem{Follin2025b}
\bibinfo{author}{Follin, L.~F.} \emph{et~al.}
\newblock \bibinfo{journal}{\bibinfo{title}{An inter-rater variability study between human and automatic scorers in 5-s mini-epochs of sleep}}.
\newblock {\emph{\JournalTitle{Sleep Med.}}} \textbf{\bibinfo{volume}{128}}, \bibinfo{pages}{139--150}, \doiprefix\url{https://doi.org/10.1016/j.sleep.2025.02.005} (\bibinfo{year}{2025}).

\bibitem{Tautan2022}
\bibinfo{author}{Tăuțan, A.-M.}, \bibinfo{author}{Rossi, A.~C.} \& \bibinfo{author}{Ionescu, B.}
\newblock \bibinfo{journal}{\bibinfo{title}{Automatic sleep scoring with LSTM networks: impact of time granularity and input signals}}.
\newblock {\emph{\JournalTitle{{Biomed. Eng. Biomed. Tech.}}}} \textbf{\bibinfo{volume}{67}}, \bibinfo{pages}{267--281}, \doiprefix\url{https://doi.org/10.1515/bmt-2021-0408} (\bibinfo{year}{2022}).

\bibitem{Follin2025}
\bibinfo{author}{Follin, L.~F.} \emph{et~al.}
\newblock \bibinfo{journal}{\bibinfo{title}{Optimizing automated sleep stage scoring of 5-s mini-epochs: a transfer learning study}}.
\newblock {\emph{\JournalTitle{Sleep}}} \doiprefix\url{https://doi.org/10.1093/sleep/zsaf393} (\bibinfo{year}{2025}).

\bibitem{Stephansen2018}
\bibinfo{author}{Stephansen, J.~B.} \emph{et~al.}
\newblock \bibinfo{journal}{\bibinfo{title}{Neural network analysis of sleep stages enables efficient diagnosis of narcolepsy}}.
\newblock {\emph{\JournalTitle{Nat. Commun.}}} \textbf{\bibinfo{volume}{9}}, \doiprefix\url{https://doi.org/10.1038/s41467-018-07229-3} (\bibinfo{year}{2018}).

\bibitem{Koch2018}
\bibinfo{author}{Koch, H.}, \bibinfo{author}{Jennum, P.} \& \bibinfo{author}{Christensen, J. A.~E.}
\newblock \bibinfo{journal}{\bibinfo{title}{Automatic sleep classification using adaptive segmentation reveals an increased number of rapid eye movement sleep transitions}}.
\newblock {\emph{\JournalTitle{J. Sleep Res.}}} \textbf{\bibinfo{volume}{28}}, \doiprefix\url{https://doi.org/10.1111/jsr.12780} (\bibinfo{year}{2018}).

\bibitem{Olesen2020}
\bibinfo{author}{Olesen, A.~N.}, \bibinfo{author}{Jennum, P.~J.}, \bibinfo{author}{Mignot, E.} \& \bibinfo{author}{Sorensen, H. B.~D.}
\newblock \bibinfo{journal}{\bibinfo{title}{Automatic sleep stage classification with deep residual networks in a mixed-cohort setting}}.
\newblock {\emph{\JournalTitle{Sleep}}} \textbf{\bibinfo{volume}{44}}, \doiprefix\url{https://doi.org/10.1093/sleep/zsaa161} (\bibinfo{year}{2020}).

\bibitem{Perslev2021}
\bibinfo{author}{Perslev, M.} \emph{et~al.}
\newblock \bibinfo{journal}{\bibinfo{title}{{U-Sleep}: {r}esilient high-frequency sleep staging}}.
\newblock {\emph{\JournalTitle{npj Digit. Med.}}} \textbf{\bibinfo{volume}{4}}, \doiprefix\url{https://doi.org/10.1038/S41746-021-00440-5} (\bibinfo{year}{2021}).

\bibitem{Zan2023}
\bibinfo{author}{Zan, H.} \& \bibinfo{author}{Yildiz, A.}
\newblock \bibinfo{journal}{\bibinfo{title}{Multi-task learning for arousal and sleep stage detection using fully convolutional networks}}.
\newblock {\emph{\JournalTitle{J. Neural Eng.}}} \textbf{\bibinfo{volume}{20}}, \bibinfo{pages}{056034}, \doiprefix\url{https://doi.org/10.1088/1741-2552/acfe3a} (\bibinfo{year}{2023}).

\bibitem{Krauss2021}
\bibinfo{author}{Krauss, P.} \emph{et~al.}
\newblock \bibinfo{journal}{\bibinfo{title}{Analysis and visualization of sleep stages based on deep neural networks}}.
\newblock {\emph{\JournalTitle{{Neurobiol. Sleep Circadian Rhythms}}}} \textbf{\bibinfo{volume}{10}}, \bibinfo{pages}{100064}, \doiprefix\url{https://doi.org/10.1016/j.nbscr.2021.100064} (\bibinfo{year}{2021}).

\bibitem{Guillot2021}
\bibinfo{author}{Guillot, A.} \& \bibinfo{author}{Thorey, V.}
\newblock \bibinfo{journal}{\bibinfo{title}{Robust{S}leep{N}et: {T}ransfer learning for automated sleep staging at scale}}.
\newblock {\emph{\JournalTitle{{IEEE} T. Neur. Sys. Reh.}}} \textbf{\bibinfo{volume}{29}}, \bibinfo{pages}{1441--1451}, \doiprefix\url{https://doi.org/10.1109/tnsre.2021.3098968} (\bibinfo{year}{2021}).

\bibitem{Vallat2021}
\bibinfo{author}{Vallat, R.} \& \bibinfo{author}{Walker, M.~P.}
\newblock \bibinfo{journal}{\bibinfo{title}{An open-source, high-performance tool for automated sleep staging}}.
\newblock {\emph{\JournalTitle{eLife}}} \textbf{\bibinfo{volume}{10}}, \bibinfo{pages}{e70092}, \doiprefix\url{https://doi.org/10.7554/elife.70092} (\bibinfo{year}{2021}).

\bibitem{Hanna2023}
\bibinfo{author}{Hanna, J.} \& \bibinfo{author}{Fl{\"{o}}el, A.}
\newblock \bibinfo{journal}{\bibinfo{title}{An accessible and versatile deep learning-based sleep stage classifier}}.
\newblock {\emph{\JournalTitle{Front. Neuroinform.}}} \textbf{\bibinfo{volume}{17}}, \doiprefix\url{https://doi.org/10.3389/FNINF.2023.1086634} (\bibinfo{year}{2023}).

\bibitem{Shi2024}
\bibinfo{author}{Shi, E.} \emph{et~al.}
\newblock \bibinfo{journal}{\bibinfo{title}{FoME: a foundation model for {EEG} using adaptive temporal-lateral attention scaling}}.
\newblock {\emph{\JournalTitle{Comput. Med. Imag. Grap.}}} \textbf{\bibinfo{volume}{135}}, \bibinfo{pages}{102817}, \doiprefix\url{https://doi.org/10.1016/j.compmedimag.2026.102817} (\bibinfo{year}{2026}).

\bibitem{Rossi2025b}
\bibinfo{author}{Rossi, A.~D.} \emph{et~al.}
\newblock \bibinfo{journal}{\bibinfo{title}{{SLEEPYLAND}: trust begins with fair evaluation of automatic sleep staging models}}.
\newblock {\emph{\JournalTitle{npj Digit. Med.}}} \textbf{\bibinfo{volume}{9}}, \doiprefix\url{https://doi.org/10.1038/s41746-025-02237-2} (\bibinfo{year}{2025}).

\bibitem{Berry2020}
\bibinfo{author}{Berry, R.~B.} \emph{et~al.}
\newblock \emph{\bibinfo{title}{The {AASM} Manual for the Scoring of Sleep and Associated Events: {R}ules, Terminology and Technical Specifications, Version 2.6}} (\bibinfo{publisher}{American Academy of Sleep Medicine}, \bibinfo{year}{2020}).

\bibitem{Chambon2018}
\bibinfo{author}{Chambon, S.}, \bibinfo{author}{Galtier, M.~N.}, \bibinfo{author}{Arnal, P.~J.}, \bibinfo{author}{Wainrib, G.} \& \bibinfo{author}{Gramfort, A.}
\newblock \bibinfo{journal}{\bibinfo{title}{A deep learning architecture for temporal sleep stage classification using multivariate and multimodal time series}}.
\newblock {\emph{\JournalTitle{{IEEE} T. Neur. Sys. Reh.}}} \textbf{\bibinfo{volume}{26}}, \bibinfo{pages}{758--769}, \doiprefix\url{https://doi.org/10.1109/tnsre.2018.2813138} (\bibinfo{year}{2018}).

\bibitem{Phan2019}
\bibinfo{author}{Phan, H.}, \bibinfo{author}{Andreotti, F.}, \bibinfo{author}{Cooray, N.}, \bibinfo{author}{Ch{\'{e}}n, O.~Y.} \& \bibinfo{author}{De~Vos, M.}
\newblock \bibinfo{journal}{\bibinfo{title}{Joint classification and prediction {CNN} framework for automatic sleep stage classification}}.
\newblock {\emph{\JournalTitle{{IEEE} Trans. Biomed. Eng.}}} \textbf{\bibinfo{volume}{66}}, \bibinfo{pages}{1285--1296}, \doiprefix\url{https://doi.org/10.1109/TBME.2018.2872652} (\bibinfo{year}{2019}).

\bibitem{Ronneberger2015}
\bibinfo{author}{Ronneberger, O.}, \bibinfo{author}{Fischer, P.} \& \bibinfo{author}{Brox, T.}
\newblock \bibinfo{title}{{U}-{N}et: Convolutional networks for biomedical image segmentation}.
\newblock In \emph{\bibinfo{booktitle}{Lecture Notes in Computer Science}}, \bibinfo{pages}{234--241}, \doiprefix\url{https://doi.org/10.1007/978-3-319-24574-4\_28} (\bibinfo{publisher}{Springer International Publishing}, \bibinfo{year}{2015}).

\bibitem{OReilly2014}
\bibinfo{author}{O'Reilly, C.}, \bibinfo{author}{Gosselin, N.}, \bibinfo{author}{Carrier, J.} \& \bibinfo{author}{Nielsen, T.}
\newblock \bibinfo{journal}{\bibinfo{title}{Montreal archive of sleep studies: {a}n open-access resource for instrument benchmarking and exploratory research}}.
\newblock {\emph{\JournalTitle{J. Sleep Res.}}} \textbf{\bibinfo{volume}{23}}, \bibinfo{pages}{628--635}, \doiprefix\url{https://doi.org/10.1111/jsr.12169} (\bibinfo{year}{2014}).

\bibitem{Hermans2022}
\bibinfo{author}{Hermans, L.~W.} \emph{et~al.}
\newblock \bibinfo{journal}{\bibinfo{title}{Representations of temporal sleep dynamics: {r}eview and synthesis of the literature}}.
\newblock {\emph{\JournalTitle{Sleep Med. Rev.}}} \textbf{\bibinfo{volume}{63}}, \bibinfo{pages}{101611}, \doiprefix\url{https://doi.org/10.1016/j.smrv.2022.101611} (\bibinfo{year}{2022}).

\bibitem{Bonnet2003}
\bibinfo{author}{Bonnet, M.~H.} \& \bibinfo{author}{Arand, D.~L.}
\newblock \bibinfo{journal}{\bibinfo{title}{Clinical effects of sleep fragmentation versus sleep deprivation}}.
\newblock {\emph{\JournalTitle{Sleep Med. Rev.}}} \textbf{\bibinfo{volume}{7}}, \bibinfo{pages}{297--310}, \doiprefix\url{https://doi.org/10.1053/smrv.2001.0245} (\bibinfo{year}{2003}).

\bibitem{Christensen2015}
\bibinfo{author}{Christensen, J. A.~E.} \emph{et~al.}
\newblock \bibinfo{journal}{\bibinfo{title}{Sleep-stage transitions during polysomnographic recordings as diagnostic features of type 1 narcolepsy}}.
\newblock {\emph{\JournalTitle{Sleep Med.}}} \textbf{\bibinfo{volume}{16}}, \bibinfo{pages}{1558--1566}, \doiprefix\url{https://doi.org/10.1016/j.sleep.2015.06.007} (\bibinfo{year}{2015}).

\bibitem{Conte2023}
\bibinfo{author}{Conte, F.} \emph{et~al.}
\newblock \bibinfo{journal}{\bibinfo{title}{Sleep continuity, stability and cyclic organization are impaired in insomniacs: {a} case-control study}}.
\newblock {\emph{\JournalTitle{{Int. J. Env. Res. Public Health}}}} \textbf{\bibinfo{volume}{20}}, \bibinfo{pages}{1240}, \doiprefix\url{https://doi.org/10.3390/ijerph20021240} (\bibinfo{year}{2023}).

\bibitem{Baglioni2014}
\bibinfo{author}{Baglioni, C.} \emph{et~al.}
\newblock \bibinfo{journal}{\bibinfo{title}{Sleep changes in the disorder of insomnia: {a} meta-analysis of polysomnographic studies}}.
\newblock {\emph{\JournalTitle{Sleep Med. Rev.}}} \textbf{\bibinfo{volume}{18}}, \bibinfo{pages}{195--213}, \doiprefix\url{https://doi.org/10.1016/j.smrv.2013.04.001} (\bibinfo{year}{2014}).

\bibitem{Waechter2019}
\bibinfo{author}{W\"{a}chter, M.} \emph{et~al.}
\newblock \bibinfo{journal}{\bibinfo{title}{Unique sleep-stage transitions determined by obstructive sleep apnea severity, age and gender}}.
\newblock {\emph{\JournalTitle{J. Sleep Res.}}} \textbf{\bibinfo{volume}{29}}, \doiprefix\url{https://doi.org/10.1111/jsr.12895} (\bibinfo{year}{2019}).

\bibitem{Bianchi2010}
\bibinfo{author}{Bianchi, M.~T.}, \bibinfo{author}{Cash, S.~S.}, \bibinfo{author}{Mietus, J.}, \bibinfo{author}{Peng, C.-K.} \& \bibinfo{author}{Thomas, R.}
\newblock \bibinfo{journal}{\bibinfo{title}{Obstructive sleep apnea alters sleep stage transition dynamics}}.
\newblock {\emph{\JournalTitle{PLoS ONE}}} \textbf{\bibinfo{volume}{5}}, \bibinfo{pages}{e11356}, \doiprefix\url{https://doi.org/10.1371/journal.pone.0011356} (\bibinfo{year}{2010}).

\bibitem{Liu2015}
\bibinfo{author}{Liu, Y.} \emph{et~al.}
\newblock \bibinfo{journal}{\bibinfo{title}{Altered sleep stage transitions of REM sleep: {a} novel and stable biomarker of narcolepsy}}.
\newblock {\emph{\JournalTitle{J. Clin. Sleep Med.}}} \textbf{\bibinfo{volume}{11}}, \bibinfo{pages}{885--894}, \doiprefix\url{https://doi.org/10.5664/jcsm.4940} (\bibinfo{year}{2015}).

\bibitem{Hermans2021}
\bibinfo{author}{Hermans, L.~W.~A.} \emph{et~al.}
\newblock \bibinfo{journal}{\bibinfo{title}{Sleep-wake survival dynamics in people with insomnia}}.
\newblock {\emph{\JournalTitle{Nat. Sci. Sleep}}} \textbf{\bibinfo{volume}{13}}, \bibinfo{pages}{349--360}, \doiprefix\url{https://doi.org/10.2147/nss.s295699} (\bibinfo{year}{2021}).

\bibitem{Wei2017}
\bibinfo{author}{Wei, Y.} \emph{et~al.}
\newblock \bibinfo{journal}{\bibinfo{title}{Sleep stage transition dynamics reveal specific stage 2 vulnerability in insomnia}}.
\newblock {\emph{\JournalTitle{Sleep}}} \doiprefix\url{https://doi.org/10.1093/sleep/zsx117} (\bibinfo{year}{2017}).

\bibitem{Stahlschmidt2022}
\bibinfo{author}{Stahlschmidt, S.~R.}, \bibinfo{author}{Ulfenborg, B.} \& \bibinfo{author}{Synnergren, J.}
\newblock \bibinfo{journal}{\bibinfo{title}{Multimodal deep learning for biomedical data fusion: a review}}.
\newblock {\emph{\JournalTitle{{Brief. Bioinform.}}}} \textbf{\bibinfo{volume}{23}}, \doiprefix\url{https://doi.org/10.1093/BIB/BBAB569} (\bibinfo{year}{2022}).

\bibitem{Rossi2025}
\bibinfo{author}{Rossi, A.~D.} \emph{et~al.}
\newblock \bibinfo{title}{{NAP}: {a}ttention-based late fusion for automatic sleep staging}. Preprint at \url{https://arxiv.org/abs/2511.03488v1} (\bibinfo{year}{2025}).

\bibitem{Phan2022}
\bibinfo{author}{Phan, H.} \& \bibinfo{author}{Mikkelsen, K.}
\newblock \bibinfo{journal}{\bibinfo{title}{Automatic sleep staging of {EEG} signals: {r}ecent development, challenges, and future directions}}.
\newblock {\emph{\JournalTitle{Physiol. Meas.}}} \textbf{\bibinfo{volume}{43}}, \bibinfo{pages}{04TR01}, \doiprefix\url{https://doi.org/10.1088/1361-6579/ac6049} (\bibinfo{year}{2022}).

\bibitem{vanGorp2022}
\bibinfo{author}{van Gorp, H.} \emph{et~al.}
\newblock \bibinfo{journal}{\bibinfo{title}{Certainty about uncertainty in sleep staging: a theoretical framework}}.
\newblock {\emph{\JournalTitle{Sleep}}} \textbf{\bibinfo{volume}{45}}, \doiprefix\url{https://doi.org/10.1093/sleep/zsac134} (\bibinfo{year}{2022}).

\bibitem{Bechny2025}
\bibinfo{author}{Bechny, M.} \emph{et~al.}
\newblock \bibinfo{journal}{\bibinfo{title}{Beyond accuracy: a framework for evaluating algorithmic bias and performance, applied to automated sleep scoring}}.
\newblock {\emph{\JournalTitle{Sci. Rep.}}} \textbf{\bibinfo{volume}{15}}, \doiprefix\url{https://doi.org/10.1038/s41598-025-06019-4} (\bibinfo{year}{2025}).

\bibitem{Chambon2018b}
\bibinfo{author}{Chambon, S.}, \bibinfo{author}{Galtier, M.~N.} \& \bibinfo{author}{Gramfort, A.}
\newblock \bibinfo{title}{Domain adaptation with optimal transport improves {EEG} sleep stage classifiers}.
\newblock In \emph{\bibinfo{booktitle}{Proc. 2018 International Workshop on Pattern Recognition in Neuroimaging, {PRNI} 2018}}, \bibinfo{pages}{1--4}, \doiprefix\url{https://doi.org/10.1109/PRNI.2018.8423957} (\bibinfo{publisher}{{IEEE}}, \bibinfo{year}{2018}).

\bibitem{Phan2021}
\bibinfo{author}{Phan, H.} \emph{et~al.}
\newblock \bibinfo{journal}{\bibinfo{title}{Towards more accurate automatic sleep staging via deep transfer learning}}.
\newblock {\emph{\JournalTitle{{IEEE} Trans. Biomed. Eng.}}} \textbf{\bibinfo{volume}{68}}, \bibinfo{pages}{1787--1798}, \doiprefix\url{https://doi.org/10.1109/TBME.2020.3020381} (\bibinfo{year}{2021}).

\bibitem{Esfahani2023}
\bibinfo{author}{Esfahani, M.~J.} \emph{et~al.}
\newblock \bibinfo{journal}{\bibinfo{title}{Validation of the sleep {EEG} headband {ZMax}}}.
\newblock Preprint at {\emph{\JournalTitle{bioRxiv}}} \doiprefix\url{https://doi.org/10.1101/2023.08.18.553744} (\bibinfo{year}{2023}).

\bibitem{Markov2025}
\bibinfo{author}{Markov, K.}, \bibinfo{author}{Elgendi, M.} \& \bibinfo{author}{Menon, C.}
\newblock \bibinfo{journal}{\bibinfo{title}{Evaluating the performance of wearable EEG sleep monitoring devices: a meta-analysis approach}}.
\newblock {\emph{\JournalTitle{npj Biomed. Innov.}}} \textbf{\bibinfo{volume}{2}}, \doiprefix\url{https://doi.org/10.1038/s44385-025-00034-w} (\bibinfo{year}{2025}).

\bibitem{Brink2025}
\bibinfo{author}{ten Brink, M.} \emph{et~al.}
\newblock \bibinfo{journal}{\bibinfo{title}{Challenges and methodological considerations for research on the role of sleep stage transitions in altered affective processing}}.
\newblock {\emph{\JournalTitle{Sleep Adv.}}} \textbf{\bibinfo{volume}{6}}, \doiprefix\url{https://doi.org/10.1093/sleepadvances/zpaf052} (\bibinfo{year}{2025}).

\bibitem{Cesari2021c}
\bibinfo{author}{Cesari, M.} \emph{et~al.}
\newblock \bibinfo{journal}{\bibinfo{title}{Sleep modelled as a continuous and dynamic process predicts healthy ageing better than traditional sleep scoring}}.
\newblock {\emph{\JournalTitle{Sleep Med.}}} \textbf{\bibinfo{volume}{77}}, \bibinfo{pages}{136--146}, \doiprefix\url{https://doi.org/10.1016/j.sleep.2020.11.033} (\bibinfo{year}{2021}).

\bibitem{Anderer2023}
\bibinfo{author}{Anderer, P.} \emph{et~al.}
\newblock \bibinfo{journal}{\bibinfo{title}{Overview of the hypnodensity approach to scoring sleep for polysomnography and home sleep testing}}.
\newblock {\emph{\JournalTitle{Front. Sleep}}} \textbf{\bibinfo{volume}{2}}, \doiprefix\url{https://doi.org/10.3389/frsle.2023.1163477} (\bibinfo{year}{2023}).

\bibitem{Bakker2018}
\bibinfo{author}{Bakker, J.~P.} \emph{et~al.}
\newblock \bibinfo{journal}{\bibinfo{title}{Gastric banding surgery versus continuous positive airway pressure for obstructive sleep apnea: {a} randomized controlled trial}}.
\newblock {\emph{\JournalTitle{Am. J. Resp. Crit. Care}}} \textbf{\bibinfo{volume}{197}}, \bibinfo{pages}{1080--1083}, \doiprefix\url{https://doi.org/10.1164/rccm.201708-1637le} (\bibinfo{year}{2018}).

\bibitem{Zhang2018}
\bibinfo{author}{Zhang, G.-Q.} \emph{et~al.}
\newblock \bibinfo{journal}{\bibinfo{title}{The national sleep research resource: {t}owards a sleep data commons}}.
\newblock {\emph{\JournalTitle{J. Am. Med. Inform. Assn.}}} \textbf{\bibinfo{volume}{25}}, \bibinfo{pages}{1351--1358}, \doiprefix\url{https://doi.org/10.1093/jamia/ocy064} (\bibinfo{year}{2018}).

\bibitem{Rosen2003}
\bibinfo{author}{Rosen, C.~L.} \emph{et~al.}
\newblock \bibinfo{journal}{\bibinfo{title}{Prevalence and risk factors for sleep-disordered breathing in 8- to 11-year-old children: {a}ssociation with race and prematurity}}.
\newblock {\emph{\JournalTitle{J. Pediatr.}}} \textbf{\bibinfo{volume}{142}}, \bibinfo{pages}{383--389}, \doiprefix\url{https://doi.org/10.1067/mpd.2003.28} (\bibinfo{year}{2003}).

\bibitem{Redline1995}
\bibinfo{author}{Redline, S.} \emph{et~al.}
\newblock \bibinfo{journal}{\bibinfo{title}{The familial aggregation of obstructive sleep apnea}}.
\newblock {\emph{\JournalTitle{Am. J. Resp. Crit. Care}}} \textbf{\bibinfo{volume}{151}}, \bibinfo{pages}{682--687}, \doiprefix\url{https://doi.org/10.1164/ajrccm/151.3\_pt\_1.682} (\bibinfo{year}{1995}).

\bibitem{Marcus2013}
\bibinfo{author}{Marcus, C.~L.} \emph{et~al.}
\newblock \bibinfo{journal}{\bibinfo{title}{A randomized trial of adenotonsillectomy for childhood sleep apnea}}.
\newblock {\emph{\JournalTitle{New Engl. J. Med.}}} \textbf{\bibinfo{volume}{368}}, \bibinfo{pages}{2366--2376}, \doiprefix\url{https://doi.org/10.1056/nejmoa1215881} (\bibinfo{year}{2013}).

\bibitem{Perslev2021c}
\bibinfo{author}{Perslev, M.} \emph{et~al.}
\newblock \bibinfo{title}{DCSM sleep staging dataset}, \doiprefix\url{https://doi.org/10.17894/UCPH.282D3C1E-9B98-4C1E-886E-704AFDFA9179} (\bibinfo{year}{2021}).

\bibitem{Rosen2012}
\bibinfo{author}{Rosen, C.~L.} \emph{et~al.}
\newblock \bibinfo{journal}{\bibinfo{title}{A multisite randomized trial of portable sleep studies and positive airway pressure autotitration versus laboratory-based polysomnography for the diagnosis and treatment of obstructive sleep apnea: {t}he {HomePAP} study}}.
\newblock {\emph{\JournalTitle{Sleep}}} \textbf{\bibinfo{volume}{35}}, \bibinfo{pages}{757--767}, \doiprefix\url{https://doi.org/10.5665/sleep.1870} (\bibinfo{year}{2012}).

\bibitem{Chen2015}
\bibinfo{author}{Chen, X.} \emph{et~al.}
\newblock \bibinfo{journal}{\bibinfo{title}{Racial/{E}thnic differences in sleep disturbances: {t}he multi-ethnic study of atherosclerosis ({MESA})}}.
\newblock {\emph{\JournalTitle{Sleep}}} \textbf{\bibinfo{volume}{38}}, \bibinfo{pages}{877--888}, \doiprefix\url{https://doi.org/10.5665/sleep.4732} (\bibinfo{year}{2015}).

\bibitem{Blackwell2011}
\bibinfo{author}{Blackwell, T.} \emph{et~al.}
\newblock \bibinfo{journal}{\bibinfo{title}{Associations between sleep architecture and sleep-disordered breathing and cognition in older community-dwelling men: {t}he osteoporotic fractures in men sleep study}}.
\newblock {\emph{\JournalTitle{J. Am. Geriatr. Soc.}}} \textbf{\bibinfo{volume}{59}}, \bibinfo{pages}{2217--2225}, \doiprefix\url{https://doi.org/10.1111/j.1532-5415.2011.03731.x} (\bibinfo{year}{2011}).

\bibitem{Ghassemi2018}
\bibinfo{author}{Ghassemi, M.~M.} \emph{et~al.}
\newblock \bibinfo{title}{You snooze, you win: {t}he PhysioNet/Computing in cardiology challenge 2018}.
\newblock In \emph{\bibinfo{booktitle}{Proc. Computing in Cardiology, {CinC} 2018}}, \bibinfo{pages}{1--4}, \doiprefix\url{https://doi.org/10.22489/CINC.2018.049} (\bibinfo{year}{2018}).

\bibitem{Goldberger2000}
\bibinfo{author}{Goldberger, A.~L.} \emph{et~al.}
\newblock \bibinfo{journal}{\bibinfo{title}{{PhysioBank}, {PhysioToolkit}, and {PhysioNet}: {c}omponents of a new research resource for complex physiologic signals}}.
\newblock {\emph{\JournalTitle{Circulation}}} \textbf{\bibinfo{volume}{101}}, \doiprefix\url{https://doi.org/10.1161/01.cir.101.23.e215} (\bibinfo{year}{2000}).

\bibitem{Kemp2000}
\bibinfo{author}{Kemp, B.}, \bibinfo{author}{Zwinderman, A.~H.}, \bibinfo{author}{Tuk, B.}, \bibinfo{author}{Kamphuisen, H. A.~C.} \& \bibinfo{author}{Oberye, J. J.~L.}
\newblock \bibinfo{journal}{\bibinfo{title}{Analysis of a sleep-dependent neuronal feedback loop: {t}he slow-wave microcontinuity of the {EEG}}}.
\newblock {\emph{\JournalTitle{{IEEE} Trans. Biomed. Eng.}}} \textbf{\bibinfo{volume}{47}}, \bibinfo{pages}{1185--1194}, \doiprefix\url{https://doi.org/10.1109/10.867928} (\bibinfo{year}{2000}).

\bibitem{Quan1997}
\bibinfo{author}{Quan, S.~F.} \emph{et~al.}
\newblock \bibinfo{journal}{\bibinfo{title}{The sleep heart health study: {d}esign, rationale, and methods}}.
\newblock {\emph{\JournalTitle{Sleep}}} \textbf{\bibinfo{volume}{20}}, \bibinfo{pages}{1077--1085}, \doiprefix\url{https://doi.org/10.1093/sleep/20.12.1077} (\bibinfo{year}{1997}).

\bibitem{Spira2007}
\bibinfo{author}{Spira, A.~P.} \emph{et~al.}
\newblock \bibinfo{journal}{\bibinfo{title}{Sleep-disordered breathing and cognition in older women}}.
\newblock {\emph{\JournalTitle{Journal of the American Geriatrics Society}}} \textbf{\bibinfo{volume}{56}}, \bibinfo{pages}{45--50}, \doiprefix\url{https://doi.org/10.1111/j.1532-5415.2007.01506.x} (\bibinfo{year}{2007}).

\bibitem{Guillot2020}
\bibinfo{author}{Guillot, A.}, \bibinfo{author}{Sauvet, F.}, \bibinfo{author}{During, E.~H.} \& \bibinfo{author}{Thorey, V.}
\newblock \bibinfo{journal}{\bibinfo{title}{Dreem open datasets: {m}ulti-scored sleep datasets to compare human and automated sleep staging}}.
\newblock {\emph{\JournalTitle{{IEEE Trans. Neur. Sys. Reh.}}}} \textbf{\bibinfo{volume}{28}}, \bibinfo{pages}{1955--1965}, \doiprefix\url{https://doi.org/10.1109/tnsre.2020.3011181} (\bibinfo{year}{2020}).

\bibitem{Khalighi2016}
\bibinfo{author}{Khalighi, S.}, \bibinfo{author}{Sousa, T.}, \bibinfo{author}{dos Santos, J.~M.} \& \bibinfo{author}{Nunes, U.}
\newblock \bibinfo{journal}{\bibinfo{title}{{ISRUC-Sleep}: {a} comprehensive public dataset for sleep researchers}}.
\newblock {\emph{\JournalTitle{Comput. Methods Programs Biomed.}}} \textbf{\bibinfo{volume}{124}}, \bibinfo{pages}{180--192}, \doiprefix\url{https://doi.org/10.1016/J.CMPB.2015.10.013} (\bibinfo{year}{2016}).

\bibitem{McNicholas2004}
\bibinfo{author}{McNicholas, W.} \emph{et~al.}
\newblock \bibinfo{title}{St. Vincent's University Hospital / University College Dublin Sleep Apnea Database}, \doiprefix\url{https://doi.org/10.13026/C26C7D} (\bibinfo{year}{2004}).

\bibitem{Andlauer2012}
\bibinfo{author}{Andlauer, O.} \emph{et~al.}
\newblock \bibinfo{journal}{\bibinfo{title}{Predictors of hypocretin (orexin) deficiency in narcolepsy without cataplexy}}.
\newblock {\emph{\JournalTitle{Sleep}}} \textbf{\bibinfo{volume}{35}}, \bibinfo{pages}{1247--1255}, \doiprefix\url{https://doi.org/10.5665/sleep.2080} (\bibinfo{year}{2012}).

\bibitem{Christensen2017}
\bibinfo{author}{Christensen, J.~A.~E.} \emph{et~al.}
\newblock \bibinfo{journal}{\bibinfo{title}{Novel method for evaluation of eye movements in patients with narcolepsy}}.
\newblock {\emph{\JournalTitle{Sleep Med.}}} \textbf{\bibinfo{volume}{33}}, \bibinfo{pages}{171--180}, \doiprefix\url{https://doi.org/10.1016/j.sleep.2016.10.016} (\bibinfo{year}{2017}).

\bibitem{Pizza2015}
\bibinfo{author}{Pizza, F.} \emph{et~al.}
\newblock \bibinfo{journal}{\bibinfo{title}{Nocturnal sleep dynamics identify narcolepsy type 1}}.
\newblock {\emph{\JournalTitle{Sleep}}} \textbf{\bibinfo{volume}{38}}, \bibinfo{pages}{1277--1284}, \doiprefix\url{https://doi.org/10.5665/sleep.4908} (\bibinfo{year}{2015}).

\bibitem{Andlauer2013}
\bibinfo{author}{Andlauer, O.} \emph{et~al.}
\newblock \bibinfo{journal}{\bibinfo{title}{Nocturnal rapid eye movement sleep latency for identifying patients with narcolepsy/hypocretin deficiency}}.
\newblock {\emph{\JournalTitle{JAMA Neurol.}}} \textbf{\bibinfo{volume}{70}}, \bibinfo{pages}{891--902}, \doiprefix\url{https://doi.org/10.1001/jamaneurol.2013.1589} (\bibinfo{year}{2013}).

\bibitem{Hong2006}
\bibinfo{author}{Hong, S.-C.} \emph{et~al.}
\newblock \bibinfo{journal}{\bibinfo{title}{A study of the diagnostic utility of {HLA} typing, {CSF} hypocretin-1 measurements, and {MSLT} testing for the diagnosis of narcolepsy in 163 {K}orean patients with unexplained excessive daytime sleepiness}}.
\newblock {\emph{\JournalTitle{Sleep}}} \textbf{\bibinfo{volume}{29}}, \bibinfo{pages}{1429--1438}, \doiprefix\url{https://doi.org/10.1093/sleep/29.11.1429} (\bibinfo{year}{2006}).

\bibitem{Moore2014}
\bibinfo{author}{Moore, H.} \emph{et~al.}
\newblock \bibinfo{journal}{\bibinfo{title}{Design and validation of a periodic leg movement detector}}.
\newblock {\emph{\JournalTitle{PLoS ONE}}} \textbf{\bibinfo{volume}{9}}, \bibinfo{pages}{e114565}, \doiprefix\url{https://doi.org/10.1371/journal.pone.0114565} (\bibinfo{year}{2014}).

\bibitem{Terzano2001}
\bibinfo{author}{Terzano, M.~G.} \emph{et~al.}
\newblock \bibinfo{journal}{\bibinfo{title}{Atlas, rules, and recording techniques for the scoring of cyclic alternating pattern ({CAP}) in human sleep}}.
\newblock {\emph{\JournalTitle{Sleep Med.}}} \textbf{\bibinfo{volume}{2}}, \bibinfo{pages}{537--553}, \doiprefix\url{https://doi.org/10.1016/s1389-9457(01)00149-6} (\bibinfo{year}{2001}).

\bibitem{Perslev2021b}
\bibinfo{author}{Perslev, M.} \emph{et~al.}
\newblock \bibinfo{title}{{U}-{S}leep data repository}, \doiprefix\url{https://doi.org/10.17894/UCPH.0D1554E9-D86B-4E08-B3C2-632B730CD362} (\bibinfo{year}{2021}).

\bibitem{Wei2024}
\bibinfo{author}{Wei, X.} \emph{et~al.}
\newblock \bibinfo{journal}{\bibinfo{title}{{ANPHY}-Sleep: an open sleep database from healthy adults using high-density scalp electroencephalogram}}.
\newblock {\emph{\JournalTitle{Sci. Data}}} \textbf{\bibinfo{volume}{11}}, \doiprefix\url{https://doi.org/10.1038/s41597-024-03722-1} (\bibinfo{year}{2024}).

\bibitem{Rechtschaffen1968}
\bibinfo{editor}{Rechtschaffen, A.} \& \bibinfo{editor}{Kales, A.} (eds.)
\newblock \emph{\bibinfo{title}{A manual of standardized terminology, techniques and scoring system for sleep stages of human subjects}} (\bibinfo{publisher}{Public Health Service, U.S. Government Printing Office}, \bibinfo{year}{1968}).

\bibitem{Vaswani2017}
\bibinfo{author}{Vaswani, A.} \emph{et~al.}
\newblock \bibinfo{title}{Attention is all you need}.
\newblock In \emph{\bibinfo{booktitle}{Proc. 30th Annual Conference on Neural Information Processing Systems, {NeurIPS}}} (eds \bibinfo{editor}{Guyon, I.} \emph{et~al.}), \bibinfo{pages}{5998--6008} (\bibinfo{publisher}{Curran Associates Inc.}, \bibinfo{year}{2017}).

\bibitem{Reddi2018}
\bibinfo{author}{Reddi, S.~J.}, \bibinfo{author}{Kale, S.} \& \bibinfo{author}{Kumar, S.}
\newblock \bibinfo{title}{On the convergence of Adam and beyond}.
\newblock In \emph{\bibinfo{booktitle}{Proc. 6th International Conference on Learning Representations, {ICLR}}} (\bibinfo{publisher}{OpenReview.net}, \bibinfo{year}{2018}).

\bibitem{Fiorillo2023}
\bibinfo{author}{Fiorillo, L.} \emph{et~al.}
\newblock \bibinfo{journal}{\bibinfo{title}{{U-Sleep}'s resilience to {AASM} guidelines}}.
\newblock {\emph{\JournalTitle{npj Digit. Med.}}} \textbf{\bibinfo{volume}{6}}, \bibinfo{pages}{33}, \doiprefix\url{https://doi.org/10.1038/S41746-023-00784-0} (\bibinfo{year}{2023}).

\bibitem{Tharwat2021}
\bibinfo{author}{Tharwat, A.}
\newblock \bibinfo{journal}{\bibinfo{title}{Classification assessment methods}}.
\newblock {\emph{\JournalTitle{Appl. Comput. Inform.}}} \textbf{\bibinfo{volume}{17}}, \bibinfo{pages}{168--192}, \doiprefix\url{https://doi.org/10.1016/j.aci.2018.08.003} (\bibinfo{year}{2021}).

\bibitem{Pedregosa2011}
\bibinfo{author}{Pedregosa, F.} \emph{et~al.}
\newblock \bibinfo{journal}{\bibinfo{title}{Scikit-learn: {m}achine learning in {P}ython}}.
\newblock {\emph{\JournalTitle{J. Mach. Learn. Res.}}} \textbf{\bibinfo{volume}{12}}, \bibinfo{pages}{2825--2830} (\bibinfo{year}{2011}).

\end{thebibliography}


\clearpage
\onecolumn
\beginappendix

\setcounter{figure}{0}
\renewcommand{\thefigure}{S\arabic{figure}}
\setcounter{table}{0}
\renewcommand{\thetable}{S\arabic{table}}
\setcounter{equation}{0}
\renewcommand{\theequation}{S\arabic{equation}}

\section{Supplementary Analyses and Validation Results}

\begin{table}[h]
  \centering
  \small
  \begin{tabular}{lr|ccccccc}
    Dataset   & $N_{\text{Rec}}$ & Wake        & N1          & N2          & N3          & REM         & MF1         & Coh. $\kappa$ \\
    \toprule
    abc       & 20               & 0.90 (.003) & 0.60 (.012) & 0.84 (.003) & 0.76 (.018) & 0.91 (.003) & 0.80 (.006) & 0.77 (.005)   \\
    ccshs     & 78               & 0.97 (.001) & 0.64 (.003) & 0.92 (.000) & 0.89 (.001) & 0.93 (.001) & 0.87 (.001) & 0.89 (.001)   \\
    cfs       & 92               & 0.96 (.001) & 0.52 (.004) & 0.89 (.000) & 0.85 (.000) & 0.91 (.001) & 0.83 (.001) & 0.86 (.001)   \\
    chat      & 128              & 0.96 (.001) & 0.65 (.004) & 0.87 (.001) & 0.91 (.000) & 0.91 (.002) & 0.86 (.001) & 0.86 (.001)   \\
    dcsm      & 39               & 0.98 (.005) & 0.52 (.008) & 0.85 (.000) & 0.82 (.002) & 0.87 (.028) & 0.81 (.005) & 0.85 (.009)   \\
    hpap      & 36               & 0.92 (.002) & 0.51 (.031) & 0.84 (.004) & 0.77 (.006) & 0.90 (.002) & 0.79 (.006) & 0.77 (.002)   \\
    mesa      & 100              & 0.95 (.001) & 0.57 (.006) & 0.86 (.001) & 0.69 (.003) & 0.91 (.001) & 0.80 (.001) & 0.82 (.001)   \\
    mros      & 134              & 0.96 (.000) & 0.45 (.006) & 0.87 (.002) & 0.71 (.002) & 0.89 (.001) & 0.78 (.001) & 0.83 (.001)   \\
    phys      & 100              & 0.84 (.006) & 0.61 (.010) & 0.84 (.002) & 0.81 (.002) & 0.88 (.002) & 0.79 (.002) & 0.74 (.001)   \\
    sedf-sc   & 23               & 0.99 (.000) & 0.58 (.005) & 0.87 (.002) & 0.73 (.017) & 0.87 (.015) & 0.81 (.004) & 0.86 (.002)   \\
    sedf-st   & 8                & 0.81 (.008) & 0.58 (.013) & 0.88 (.008) & 0.67 (.028) & 0.93 (.005) & 0.77 (.004) & 0.75 (.006)   \\
    shhs      & 140              & 0.95 (.000) & 0.50 (.007) & 0.87 (.001) & 0.78 (.002) & 0.91 (.000) & 0.80 (.002) & 0.83 (.001)   \\
    sof       & 68               & 0.96 (.001) & 0.46 (.008) & 0.86 (.002) & 0.77 (.008) & 0.92 (.002) & 0.79 (.004) & 0.83 (.003)   \\
    Mean      &                  & 0.944       & 0.548       & 0.869       & 0.794       & 0.903       & 0.812       & 0.829         \\
    \midrule
    dodh      & 25               & 0.90 (.005) & 0.63 (.022) & 0.87 (.017) & 0.80 (.022) & 0.94 (.004) & 0.83 (.012) & 0.80 (.017)   \\
    dodo      & 55               & 0.90 (.000) & 0.53 (.016) & 0.86 (.013) & 0.75 (.022) & 0.92 (.002) & 0.79 (.008) & 0.77 (.012)   \\
    isruc-sg1 & 100              & 0.89 (.001) & 0.51 (.015) & 0.80 (.001) & 0.80 (.002) & 0.88 (.001) & 0.78 (.003) & 0.73 (.002)   \\
    isruc-sg2 & 16               & 0.83 (.005) & 0.49 (.009) & 0.76 (.001) & 0.81 (.001) & 0.83 (.002) & 0.74 (.001) & 0.67 (.001)   \\
    isruc-sg3 & 10               & 0.87 (.010) & 0.53 (.011) & 0.78 (.005) & 0.76 (.017) & 0.84 (.010) & 0.76 (.007) & 0.71 (.009)   \\
    mass-c1   & 53               & 0.94 (.000) & 0.45 (.049) & 0.81 (.005) & 0.60 (.001) & 0.89 (.003) & 0.74 (.010) & 0.72 (.007)   \\
    mass-c3   & 62               & 0.92 (.002) & 0.57 (.034) & 0.86 (.010) & 0.74 (.014) & 0.91 (.003) & 0.80 (.010) & 0.77 (.011)   \\
    svuh      & 25               & 0.81 (.001) & 0.33 (.012) & 0.81 (.003) & 0.85 (.003) & 0.87 (.003) & 0.74 (.003) & 0.70 (.002)   \\
    Mean      &                  & 0.898       & 0.512       & 0.825       & 0.754       & 0.895       & 0.777       & 0.744         \\
  \end{tabular}
  \caption{
    Detailed performance overview of AnySleep on the in-distribution (upper part of the table) and hold-out test sets (lower part).
    For each dataset, scores were calculated using all available channels and then weighted by the number of test recordings to obtain weighted mean scores.
    The model was trained three times with different random seeds, and we report the mean scores and standard deviations (in parentheses).
  }
  \label{sup-tab:micros-anysleep}
\end{table}

\begin{table}[h]
  \centering
  \small
  \begin{tabular}{lr|ccccccc}
    Dataset   & $N_{\text{Rec}}$ & Wake        & N1          & N2          & N3          & REM         & MF1         & Coh. $\kappa$ \\
    \toprule
    abc       & 20               & 0.88 (.007) & 0.53 (.022) & 0.83 (.002) & 0.68 (.035) & 0.89 (.009) & 0.76 (.009) & 0.74 (.008)   \\
    ccshs     & 78               & 0.97 (.001) & 0.61 (.008) & 0.91 (.002) & 0.88 (.001) & 0.92 (.003) & 0.86 (.003) & 0.88 (.003)   \\
    cfs       & 92               & 0.96 (.002) & 0.51 (.007) & 0.88 (.002) & 0.84 (.004) & 0.90 (.006) & 0.82 (.004) & 0.85 (.003)   \\
    chat      & 128              & 0.96 (.003) & 0.60 (.023) & 0.86 (.004) & 0.90 (.001) & 0.89 (.009) & 0.84 (.007) & 0.84 (.006)   \\
    dcsm      & 39               & 0.98 (.001) & 0.50 (.016) & 0.85 (.003) & 0.80 (.011) & 0.89 (.009) & 0.81 (.005) & 0.86 (.004)   \\
    hpap      & 36               & 0.91 (.002) & 0.43 (.009) & 0.84 (.003) & 0.77 (.003) & 0.90 (.004) & 0.77 (.002) & 0.76 (.002)   \\
    mesa      & 100              & 0.95 (.001) & 0.53 (.024) & 0.86 (.001) & 0.69 (.008) & 0.90 (.002) & 0.78 (.006) & 0.81 (.003)   \\
    mros      & 134              & 0.95 (.002) & 0.43 (.011) & 0.86 (.004) & 0.69 (.013) & 0.88 (.005) & 0.76 (.007) & 0.82 (.005)   \\
    phys      & 100              & 0.84 (.004) & 0.57 (.017) & 0.84 (.003) & 0.81 (.001) & 0.88 (.002) & 0.79 (.005) & 0.74 (.004)   \\
    sedf-sc   & 23               & 0.98 (.001) & 0.56 (.011) & 0.86 (.003) & 0.75 (.006) & 0.86 (.004) & 0.80 (.003) & 0.84 (.002)   \\
    sedf-st   & 8                & 0.83 (.011) & 0.58 (.013) & 0.88 (.008) & 0.63 (.003) & 0.91 (.009) & 0.77 (.005) & 0.74 (.014)   \\
    shhs      & 140              & 0.94 (.005) & 0.48 (.010) & 0.86 (.003) & 0.76 (.009) & 0.91 (.002) & 0.79 (.005) & 0.81 (.006)   \\
    sof       & 68               & 0.96 (.001) & 0.45 (.015) & 0.85 (.005) & 0.73 (.017) & 0.91 (.003) & 0.78 (.007) & 0.82 (.006)   \\
    Mean      &                  & 0.938       & 0.517       & 0.862       & 0.781       & 0.896       & 0.799       & 0.817         \\
    \midrule
    dodh      & 25               & 0.88 (.010) & 0.60 (.019) & 0.86 (.014) & 0.79 (.021) & 0.93 (.002) & 0.81 (.012) & 0.79 (.018)   \\
    dodo      & 55               & 0.89 (.014) & 0.51 (.012) & 0.87 (.005) & 0.77 (.008) & 0.93 (.004) & 0.79 (.007) & 0.78 (.009)   \\
    isruc-sg1 & 100              & 0.90 (.006) & 0.49 (.007) & 0.79 (.003) & 0.80 (.012) & 0.89 (.001) & 0.77 (.002) & 0.73 (.004)   \\
    isruc-sg2 & 16               & 0.83 (.002) & 0.46 (.005) & 0.76 (.002) & 0.80 (.011) & 0.83 (.001) & 0.74 (.002) & 0.66 (.005)   \\
    isruc-sg3 & 10               & 0.89 (.007) & 0.56 (.008) & 0.78 (.004) & 0.77 (.013) & 0.86 (.002) & 0.77 (.002) & 0.72 (.004)   \\
    mass-c1   & 53               & 0.94 (.002) & 0.36 (.027) & 0.80 (.007) & 0.60 (.013) & 0.89 (.001) & 0.72 (.009) & 0.70 (.007)   \\
    mass-c3   & 62               & 0.93 (.001) & 0.48 (.018) & 0.85 (.004) & 0.73 (.006) & 0.91 (.001) & 0.78 (.005) & 0.75 (.005)   \\
    svuh      & 25               & 0.82 (.002) & 0.34 (.011) & 0.81 (.001) & 0.84 (.024) & 0.88 (.002) & 0.74 (.004) & 0.70 (.006)   \\
    Mean      &                  & 0.896       & 0.469       & 0.822       & 0.754       & 0.898       & 0.768       & 0.738         \\
  \end{tabular}
  \caption{
    Detailed performance overview of \USleep{} on the in-distribution (upper part of the table) and hold-out test sets (lower part).
    For each dataset, scores were calculated using all available channels and then weighted by the number of test recordings to obtain weighted mean scores.
    Following Perslev~et~al.~\cite{Perslev2021}, we generated predictions for all EEG-EOG channel pairs and combined them by majority voting.
    The model was trained three times with different random seeds, and we report the mean scores and standard deviations (in parentheses).
  }
  \label{sup-tab:micros-usleep}
\end{table}

\begin{figure}[h]
  \centering
  \includegraphics[width=\linewidth]{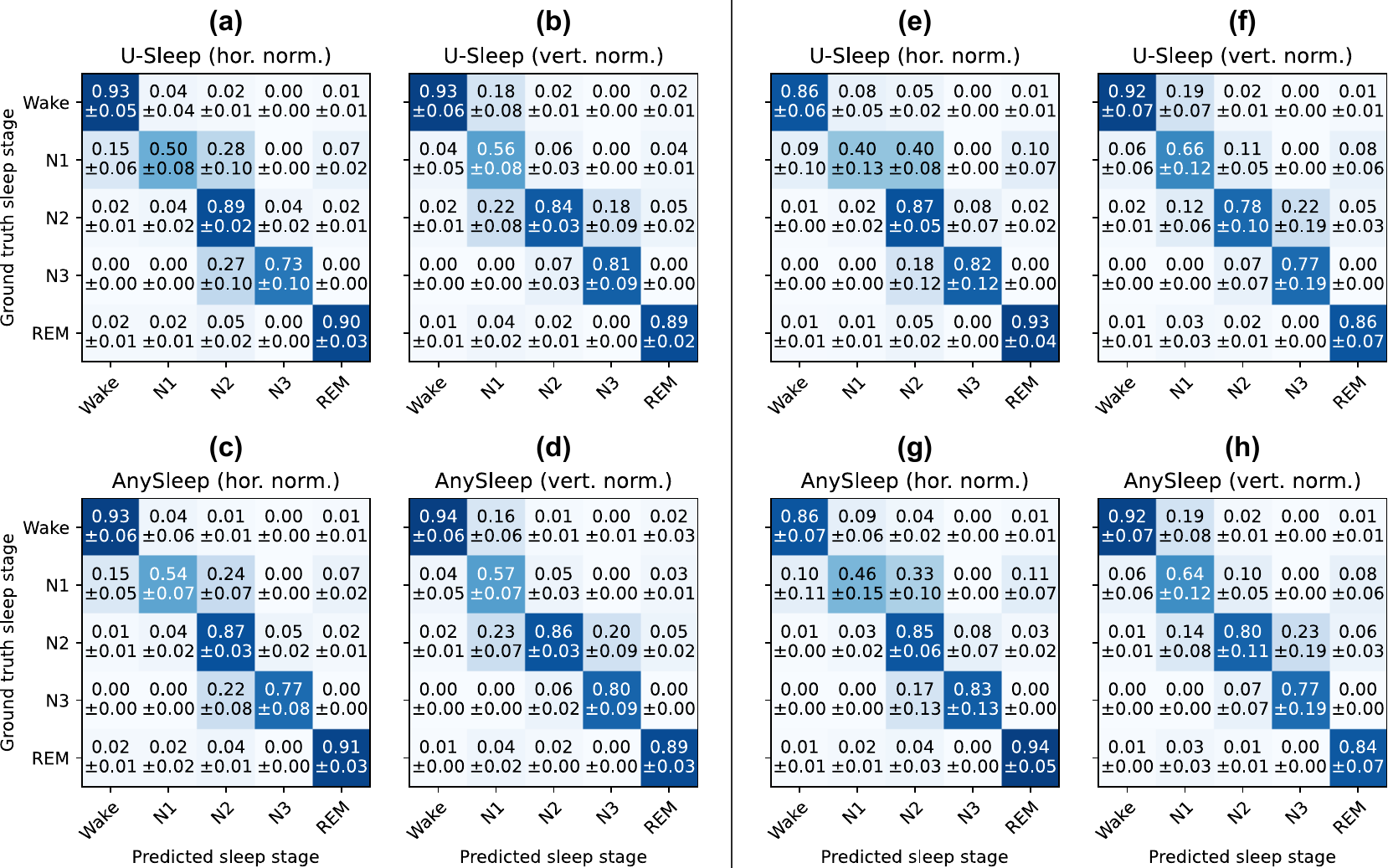}
  \caption{
    Confusion matrices for \USleep{} (upper panels \textbf{(a)}, \textbf{(b)}, \textbf{(e)}, \textbf{(f)}) and AnySleep (lower panels \textbf{(c)}, \textbf{(d)}, \textbf{(g)}, \textbf{(h)}) on the test sets.
    The left panels (\textbf{(a)}--\textbf{(d)}) show confusion matrices for the in-distribution test sets, while the right panels (\textbf{(e)}--\textbf{(h)}) show confusion matrices for the hold-out test sets.
    Matrices were calculated for each dataset and model run, normalized horizontally (panels \textbf{(a)}, \textbf{(c)}, \textbf{(e)}, \textbf{(g)}) or vertically (panels \textbf{(b)}, \textbf{(d)}, \textbf{(f)}, \textbf{(h)}), and finally averaged ($\pm$ standard deviation).
  }
  \label{sup-fig:confusion-matrices}
\end{figure}

\newcolumntype{L}{>{\raggedright\arraybackslash}p{0.09\textwidth}}
\newcolumntype{Y}{>{\raggedright\arraybackslash}p{0.62\textwidth}}
\newcolumntype{Z}{>{\raggedright\arraybackslash}p{0.20\textwidth}}
\clearpage
\FloatBarrier
\begin{longtable}{L | Y | Z}
  \\
  Dataset   & EEG channels                                                                                                                                                                                                                                                                                                                                                                                                                                                                                                                                                                                                                                                                                                                                                                                                                                                                                                                                                                                                                                                                               & EOG channels                                                                                                                                                           \\
  \toprule
  \multicolumn{3}{c}{\emph{In-Distribution training and test sets}}                                                                                                                                                                                                                                                                                                                                                                                                                                                                                                                                                                                                                                                                                                                                                                                                                                                                                                                                                                                                                                                                                                                                                                                                               \\
  \midrule
  abc       & F3-M2 (100\%), F4-M1 (100\%), C3-M2 (100\%), C4-M1 (100\%), O1-M2 (100\%), O2-M1 (100\%)                                                                                                                                                                                                                                                                                                                                                                                                                                                                                                                                                                                                                                                                                                                                                                                                                                                                                                                                                                                                   & E1-M2 (100\%), E2-M1 (100\%)                                                                                                                                           \\
  \midrule
  ccshs     & C3-A2 (100\%), C4-A1 (100\%)                                                                                                                                                                                                                                                                                                                                                                                                                                                                                                                                                                                                                                                                                                                                                                                                                                                                                                                                                                                                                                                               & LOC-A2 (100\%), ROC-A1 (100\%)                                                                                                                                         \\
  \midrule
  cfs       & C3-A2 (100\%), C4-A1 (100\%)                                                                                                                                                                                                                                                                                                                                                                                                                                                                                                                                                                                                                                                                                                                                                                                                                                                                                                                                                                                                                                                               & LOC-A2 (100\%), ROC-A1 (100\%)                                                                                                                                                         \\
  \midrule
  chat      & F3-M2 (100\%), F4-M1 (100\%), C3-M2 (100\%), C4-M1 (100\%), T3-M2 (100\%), T4-M1 (100\%), O1-M2 (100\%), O2-M1 (100\%)                                                                                                                                                                                                                                                                                                                                                                                                                                                                                                                                                                                                                                                                                                                                                                                                                                                                                                                                                                     & E1-M2 (100\%), E2-M1 (100\%)                                                                                                                                           \\
  \midrule
  dcsm      & F3-M2 (100\%), F4-M1 (100\%), C3-M2 (100\%), C4-M1 (100\%), O1-M2 (100\%), O2-M1 (100\%)                                                                                                                                                                                                                                                                                                                                                                                                                                                                                                                                                                                                                                                                                                                                                                                                                                                                                                                                                                                                   & E1-M2 (100\%), E2-M2 (100\%)                                                                                                                                           \\
  \midrule
  hpap      & F3-M2 (100\%), F4-M1 (100\%), C3-M2 (100\%), C4-M1 (100\%), O1-M2 (100\%), O2-M1 (100\%)                                                                                                                                                                                                                                                                                                                                                                                                                                                                                                                                                                                                                                                                                                                                                                                                                                                                                                                                                                                                   & E1-M2 (84\%), E2-M1 (85\%), E1 (91\%), E2 (91\%)                                                                                                                       \\
  \midrule
  mesa      & EEG1 [Fz-Cz] (100\%), EEG2 [Cz-Oz] (100\%), EEG3 [C4-M1] (100\%)                                                                                                                                                                                                                                                                                                                                                                                                                                                                                                                                                                                                                                                                                                                                                                                                                                                                                                                                                                                                                           & EOG-L (100\%), EOG-R (100\%)                                                                                                                                           \\
  \midrule
  mros      & C3-M2 (100\%), C4-M1 (100\%)                                                                                                                                                                                                                                                                                                                                                                                                                                                                                                                                                                                                                                                                                                                                                                                                                                                                                                                                                                                                                                                               & E1-M2 (100\%), E2-M1 (100\%)                                                                                                                                           \\
  \midrule
  phys      & F3-M2 (100\%), F4-M1 (100\%), C3-M2 (100\%), C4-M1 (100\%), O1-M2 (100\%), O2-M1 (100\%)                                                                                                                                                                                                                                                                                                                                                                                                                                                                                                                                                                                                                                                                                                                                                                                                                                                                                                                                                                                                   & E1-M2 (100\%)                                                                                                                                                          \\
  \midrule
  sedf-sc   & EEG Fpz-Cz (100\%), EEG Pz-Oz (100\%)                                                                                                                                                                                                                                                                                                                                                                                                                                                                                                                                                                                                                                                                                                                                                                                                                                                                                                                                                                                                                                                      & EOG horizontal (100\%)                                                                                                                                                 \\
  \midrule
  sedf-st   & EEG Fpz-Cz (100\%), EEG Pz-Oz (100\%)                                                                                                                                                                                                                                                                                                                                                                                                                                                                                                                                                                                                                                                                                                                                                                                                                                                                                                                                                                                                                                                      & EOG horizontal (100\%)                                                                                                                                                 \\
  \midrule
  shhs      & EEG [C4-A1] (100\%), EEG(sec) [C3-A2] (98\%)                                                                                                                                                                                                                                                                                                                                                                                                                                                                                                                                                                                                                                                                                                                                                                                                                                                                                                                                                                                                                                               & EOG(L) (100\%), EOG(R) (100\%)                                                                                                                                         \\
  \midrule
  sof       & C3-A2 (100\%), C4-A1 (100\%)                                                                                                                                                                                                                                                                                                                                                                                                                                                                                                                                                                                                                                                                                                                                                                                                                                                                                                                                                                                                                                                               & LOC-A2 (100\%), ROC-A1 (100\%)                                                                                                                                         \\
  \midrule
  \multicolumn{3}{c}{\emph{Hold-Out test sets}}                                                                                                                                                                                                                                                                                                                                                                                                                                                                                                                                                                                                                                                                                                                                                                                                                                                                                                                                                                                                                                                                                                                                                                                                                                   \\
  \midrule
  dodh      & C3\_M2 (100\%), F4\_M1 (100\%), F3\_F4 (100\%), F3\_M2 (100\%), F4\_O2 (100\%), F3\_O1 (100\%), FP1\_F3 (100\%), FP1\_M2 (100\%), FP1\_O1 (100\%), FP2\_F4 (100\%), FP2\_M1 (100\%), FP2\_O2 (100\%)                                                                                                                                                                                                                                                                                                                                                                                                                                                                                                                                                                                                                                                                                                                                                                                                                                                                                       & EOG1 (100\%), EOG2 (100\%)                                                                                                                                             \\
  \midrule
  dodo      & C4\_M1 (100\%), C3\_M2 (100\%), F3\_F4 (100\%), F3\_M2 (100\%), F4\_O2 (100\%), F3\_O1 (100\%), O1\_M2 (100\%), O2\_M1 (100\%)                                                                                                                                                                                                                                                                                                                                                                                                                                                                                                                                                                                                                                                                                                                                                                                                                                                                                                                                                             & EOG1 (100\%), EOG2 (100\%)                                                                                                                                             \\
  \midrule
  isruc-sg1 & F3-M2 (100\%), F4-M1 (100\%), C3-M2 (100\%), C4-M1 (100\%), O1-M2 (100\%), O2-M1 (100\%)                                                                                                                                                                                                                                                                                                                                                                                                                                                                                                                                                                                                                                                                                                                                                                                                                                                                                                                                                                                                   & E1-M2 (100\%), E2-M1 (100\%)                                                                                                                                           \\
  \midrule
  isruc-sg2 & F3-M2 (100\%), F4-M1 (100\%), C3-M2 (100\%), C4-M1 (100\%), O1-M2 (100\%), O2-M1 (100\%)                                                                                                                                                                                                                                                                                                                                                                                                                                                                                                                                                                                                                                                                                                                                                                                                                                                                                                                                                                                                   & E1-M2 (100\%), E2-M1 (100\%)                                                                                                                                           \\
  \midrule
  isruc-sg3 & F3-M2 (100\%), F4-M1 (100\%), C3-M2 (100\%), C4-M1 (100\%), O1-M2 (100\%), O2-M1 (100\%)                                                                                                                                                                                                                                                                                                                                                                                                                                                                                                                                                                                                                                                                                                                                                                                                                                                                                                                                                                                                   & E1-M2 (100\%), E2-M1 (100\%)                                                                                                                                           \\
  \midrule
  mass-c1   & EEG F3-CLE (89\%), EEG F4-CLE (89\%), EEG C3-CLE (89\%), EEG C4-CLE (89\%), EEG O1-CLE (89\%), EEG O2-CLE (89\%), EEG F7-CLE (89\%), EEG F8-CLE (89\%), EEG T3-CLE (89\%), EEG T4-CLE (89\%), EEG T5-CLE (89\%), EEG T6-CLE (89\%), EEG P3-CLE (89\%), EEG P4-CLE (89\%), EEG Fz-CLE (89\%), EEG Cz-CLE (89\%), EEG Pz-CLE (89\%), EEG F3-LER (11\%), EEG F4-LER (11\%), EEG C3-LER (11\%), EEG C4-LER (11\%), EEG O1-LER (11\%), EEG O2-LER (11\%), EEG F7-LER (11\%), EEG F8-LER (11\%), EEG T3-LER (11\%), EEG T4-LER (11\%), EEG T5-LER (11\%), EEG T6-LER (11\%), EEG P3-LER (11\%), EEG P4-LER (11\%), EEG Fz-LER (11\%), EEG Cz-LER (11\%), EEG Pz-LER (11\%), EEG Fp1-LER (6\%), EEG Fp2-LER (6\%)                                                                                                                                                                                                                                                                                                                                                                                 & EOG Left Horiz (100\%), EOG Right Horiz (100\%)                                                                                                                        \\
  \midrule
  mass-c3   & EEG Fp1-LER (100\%), EEG Fp2-LER (100\%), EEG F7-LER (100\%), EEG F8-LER (100\%), EEG F3-LER (100\%), EEG F4-LER (100\%), EEG T3-LER (100\%), EEG T4-LER (100\%), EEG C3-LER (100\%), EEG C4-LER (100\%), EEG T5-LER (100\%), EEG T6-LER (100\%), EEG P3-LER (100\%), EEG P4-LER (100\%), EEG O1-LER (100\%), EEG O2-LER (100\%), EEG Fz-LER (100\%), EEG Cz-LER (100\%), EEG Pz-LER (100\%), EEG Oz-LER (100\%), EEG A2-LER (69\%)                                                                                                                                                                                                                                                                                                                                                                                                                                                                                                                                                                                                                                                        & EOG Left Horiz (100\%), EOG Right Horiz (100\%)                                                                                                                        \\
  \midrule
  svuh      & C3A2 (100\%), C4A1 (100\%)                                                                                                                                                                                                                                                                                                                                                                                                                                                                                                                                                                                                                                                                                                                                                                                                                                                                                                                                                                                                                                                                 & Lefteye (100\%), RightEye (100\%)                                                                                                                                      \\
  \midrule
  \multicolumn{3}{c}{\emph{Test sets used for high-frequency evaluations}}                                                                                                                                                                                                                                                                                                                                                                                                                                                                                                                                                                                                                                                                                                                                                                                                                                                                                                                                                                                                                                                                                                                                                                                                        \\
  \midrule
  cnc       & C3-A2 (67\%), C3-M2 (23\%), C4-A1 (67\%), C4-M1 (23\%), EEG A1-A2 (5\%), EEG C3-A2 (10\%), EEG C4-A1 (10\%), EEG F3-A2 (10\%), EEG F4-A1 (10\%), EEG Fp2-A1 (3\%), EEG O1-A2 (10\%), EEG O2-A1 (10\%), F3-A2 (67\%), F3-M2 (23\%), F4-A1 (67\%), F4-M1 (23\%), O1-A2 (67\%), O1-M2 (23\%), O1-O2 (1\%), O2-A1 (67\%), O2-M1 (23\%), O2-O1 (1\%)                                                                                                                                                                                                                                                                                                                                                                                                                                                                                                                                                                                                                                                                                                                                            & E1-M2 (23\%), E2-M2 (23\%), EOG LOC-A2 (10\%), EOG ROC-A2 (10\%), LOC-A2 (67\%), ROC-A1 (67\%)                                                                         \\
  \midrule
  dhc       & c3a2 (100\%), c4a1 (100\%), f3a2 (100\%), f4a1 (100\%), o1a2 (98\%), o2a1 (100\%)                                                                                                                                                                                                                                                                                                                                                                                                                                                                                                                                                                                                                                                                                                                                                                                                                                                                                                                                                                                                          & eogla2 (100\%), eogra1 (100\%)                                                                                                                                         \\
  \midrule
  fhc       & EEG A1 (62\%), EEG A2 (62\%), EEG C3 (100\%), EEG C4 (100\%), EEG Fp1 (38\%), EEG Fp2 (38\%), EEG O1 (38\%), EEG O2 (99\%), EEG T3 (38\%), EEG T4 (38\%), T1 (1\%), T2 (1\%)                                                                                                                                                                                                                                                                                                                                                                                                                                                                                                                                                                                                                                                                                                                                                                                                                                                                                                               & EOG3 (100\%), EOGD (100\%), EOGG (100\%)                                                                                                                               \\
  \midrule
  ihc       & A1 (95\%), A2 (95\%), C3 (93\%), C3/A2 (1\%), C4 (95\%), C4/A1 (1\%), C4\_A1 (5\%), CZ/A1 (1\%), Cz (95\%), Cz\_A1 (2\%), O1 (95\%), O1/A2 (1\%), O1\_A2 (5\%), O2\_A1 (3\%)                                                                                                                                                                                                                                                                                                                                                                                                                                                                                                                                                                                                                                                                                                                                                                                                                                                                                                               & EOG\_L (100\%), EOG\_R (100\%)                                                                                                                                         \\
  \midrule
  khc       & A1 (71\%), A2 (71\%), C3 (71\%), C3-A2 (29\%), C4 (71\%), C4-A1 (29\%), O1 (71\%), O1-A2 (29\%), O2 (71\%), O2-A1 (29\%)                                                                                                                                                                                                                                                                                                                                                                                                                                                                                                                                                                                                                                                                                                                                                                                                                                                                                                                                                                   & EOG Left (71\%), EOG Right (71\%), EOG-L (29\%), EOG-R (29\%)                                                                                                          \\
  \midrule
  ssc       & C3-A1 (24\%), C3-A2 (99\%), C3-O1 (1\%), C4-A1 (99\%), C4-A2 (24\%), F1-A2 (96\%), F2-C4 (24\%), F2-T4 (2\%), F3-A2 (1\%), F4-A1 (1\%), FP-? (1\%), FP1-C3 (1\%), FP2-C4 (1\%), Fz-A1 (24\%), Fz-A2 (96\%), O1-x (99\%), O2-x (99\%), T3 (1\%), T3-O1 (1\%), T4-O2 (1\%)                                                                                                                                                                                                                                                                                                                                                                                                                                                                                                                                                                                                                                                                                                                                                                                                                   & L-EOG (2\%), LOC (1\%), LOC-A2 (97\%), R-EOG (2\%), ROC (1\%), ROC-A1 (97\%)                                                                                           \\
  \midrule
  cap       & A1 (2\%), A2 (2\%), C3 (2\%), C3-A2 (5\%), C3-P3 (65\%), C3A2 (3\%), C4 (2\%), C4-A1 (92\%), C4-P4 (92\%), C4A1 (3\%), F1-F3 (1\%), F2-F4 (3\%), F3 (2\%), F3-C3 (65\%), F3A2 (3\%), F4 (2\%), F4-C4 (92\%), F4A1 (3\%), F7 (2\%), F7-T3 (63\%), F8 (2\%), F8-T4 (63\%), FP1 (2\%), FP1-F3 (64\%), Fp2 (2\%), Fp2-F4 (81\%), O1 (2\%), O1-A2 (1\%), O1A2 (3\%), O2 (2\%), O2-A1 (4\%), O2A1 (3\%), P3 (2\%), P3-O1 (65\%), P4 (2\%), P4-O2 (92\%), T3 (2\%), T3-T5 (42\%), T4 (3\%), T4-T6 (42\%), T5 (2\%), T6 (2\%)                                                                                                                                                                                                                                                                                                                                                                                                                                                                                                                                                                      & EOG dx (1\%), EOG sin (1\%), EOG-L (2\%), EOG-R (2\%), LOC (4\%), LOC / A2 (1\%), LOC-A1 (3\%), LOC-ROC (3\%), ROC (4\%), ROC / A1 (1\%), ROC-A2 (3\%), ROC-LOC (89\%) \\
  \midrule
  \multicolumn{3}{c}{\emph{High-density EEG test sets}}                                                                                                                                                                                                                                                                                                                                                                                                                                                                                                                                                                                                                                                                                                                                                                                                                                                                                                                                                                                                                                                                                                                                                                                                                           \\
  \midrule
  anphy     & Fp1 (100\%), Fp2 (100\%), F3 (100\%), F4 (100\%), C3 (100\%), C4 (100\%), P3 (100\%), P4 (100\%), O1 (100\%), O2 (100\%), F7 (100\%), F8 (100\%), T3 (100\%), T4 (100\%), T5 (100\%), T6 (100\%), FZ (100\%), CZ (100\%), PZ (100\%), SO1 (100\%), SO2 (100\%), F9 (100\%), F10 (100\%), ZY1 (100\%), ZY2 (100\%), T9 (100\%), T10 (100\%), P9 (100\%), P10 (100\%), AF7 (100\%), AF3 (100\%), F11 (100\%), F5 (100\%), F1 (100\%), FT11 (100\%), FT9 (100\%), FT7 (100\%), FC5 (100\%), FC3 (100\%), FC1 (100\%), FCZ (100\%), C5 (100\%), C1 (100\%), TP11 (100\%), TP9 (100\%), TP7 (100\%), CP3 (100\%), CP1 (100\%), P11 (100\%), P5 (100\%), P1 (100\%), PO7 (100\%), PO3 (100\%), POZ (100\%), OZ (100\%), FPZ (100\%), AFZ (100\%), AF4 (100\%), AF8 (100\%), F2 (100\%), F6 (100\%), F12 (100\%), FC2 (100\%), FC4 (100\%), FC6 (100\%), FT8 (100\%), FT10 (100\%), FT12 (100\%), C6 (100\%), C2 (100\%), CPZ (100\%), CP2 (100\%), CP4 (100\%), CP6 (100\%), TP8 (100\%), TP10 (100\%), TP12 (100\%), P2 (100\%), P6 (100\%), P12 (100\%), PO4 (100\%), PO8 (100\%), CP5 (100\%)
            & EOG1 (100\%), EOG2 (100\%)                                                                                                                                                                                                                                                                                                                                                                                                                                                                                                                                                                                                                                                                                                                                                                                                                                                                                                                                                                                                                                                                                                                                                                                                                                          \\
  \caption{
    EEG and EOG channels available in each dataset.
    Datasets are grouped by their role in the study.
    Percentages refer to the proportion of recordings in which each channel was available.
    Where not self-evident, electrode configurations are given in square brackets (LER = linked-ear reference, CLE = computed linked-ear~\cite{OReilly2014}).
  }
  \label{sup-tab:channels}
\end{longtable}

\FloatBarrier

\begin{table}[h]
  \centering
  \begin{tabular}{ll|llr|ll}
    Metric           & Expert            & AnySleep          & $\Delta$ to expert & $\rho_{\text{Spear}}$ & \USleep{}         & $\Delta$ to expert \\
    \toprule
    TST              & 6:17$\pm$1:14     & 6:23$\pm$1:13     & -0:06$\pm$0:19     & 0.97                  & 6:23$\pm$1:13     & -0:06$\pm$0:18     \\
    Wake duration    & 1:30$\pm$0:59     & 1:24$\pm$0:57     & 0:06$\pm$0:19      & 0.95                  & 1:23$\pm$0:56     & 0:06$\pm$0:18      \\
    N1 duration      & 0:50$\pm$0:29     & 0:35$\pm$0:25     & 0:15$\pm$0:29      & 0.50                  & 0:29$\pm$0:21     & 0:21$\pm$0:28      \\
    N2 duration      & 3:12$\pm$1:07     & 3:18$\pm$0:49     & -0:05$\pm$0:47     & 0.69                  & 3:27$\pm$0:50     & -0:15$\pm$0:46     \\
    N3 duration      & 1:05$\pm$0:40     & 1:13$\pm$0:36     & -0:08$\pm$0:36     & 0.55                  & 1:12$\pm$0:37     & -0:07$\pm$0:36     \\
    REM duration     & 1:09$\pm$0:32     & 1:16$\pm$0:34     & -0:07$\pm$0:13     & 0.93                  & 1:14$\pm$0:34     & -0:05$\pm$0:13     \\
    \midrule
    Sleep efficiency & 78.80$\pm$14.06\% & 82.38$\pm$13.27\% & -3.57$\pm$3.57\%   & 0.96                  & 81.50$\pm$13.38\% & -2.70$\pm$3.47\%   \\
    WASO             & 1:16$\pm$0:50     & 1:01$\pm$0:47     & 0:15$\pm$0:21      & 0.96                  & 1:04$\pm$0:46     & 0:12$\pm$0:20      \\
    REM latency      & 2:06$\pm$1:01     & 2:05$\pm$1:07     & 0:00$\pm$0:39      & 0.87                  & 2:11$\pm$1:09     & -0:05$\pm$0:43     \\
    Sleep latency    & 0:18$\pm$0:20     & 0:17$\pm$0:24     & 0:01$\pm$0:19      & 0.91                  & 0:18$\pm$0:24     & 0:00$\pm$0:17     \\
  \end{tabular}
  \caption{
    Sleep metrics derived from expert annotations and predicted hypnograms from AnySleep and \USleep{} for the hold-out datasets at 30-s resolution.
    Metrics are reported as mean$\pm$standard deviation of hours:minutes across recordings, with $\Delta$ values and Spearman's $\rho$ quantifying recording-wise differences (calculated as expert$-$model) and associations between expert-derived and prediction-derived metrics, respectively.
    Total sleep time (TST) and sleep-specific durations were obtained by summing the duration of the respective stages.
    Sleep efficiency, wake after sleep onset (WASO), REM sleep latency, and sleep onset latency were calculated solely for the ISRUC datasets, the only hold-out datasets with available lights-off/on times required by the AASM~\cite{Berry2020} guidelines.
    We note that the ISRUC datasets consist of both healthy controls and patients with various disorders including sleep apnea, REM sleep disorder, and affective disorders.
  }
  \label{sup-tab:hyp-derived-metrics}
\end{table}

\begin{figure}[h]
  \centering
  \includegraphics[width=\linewidth]{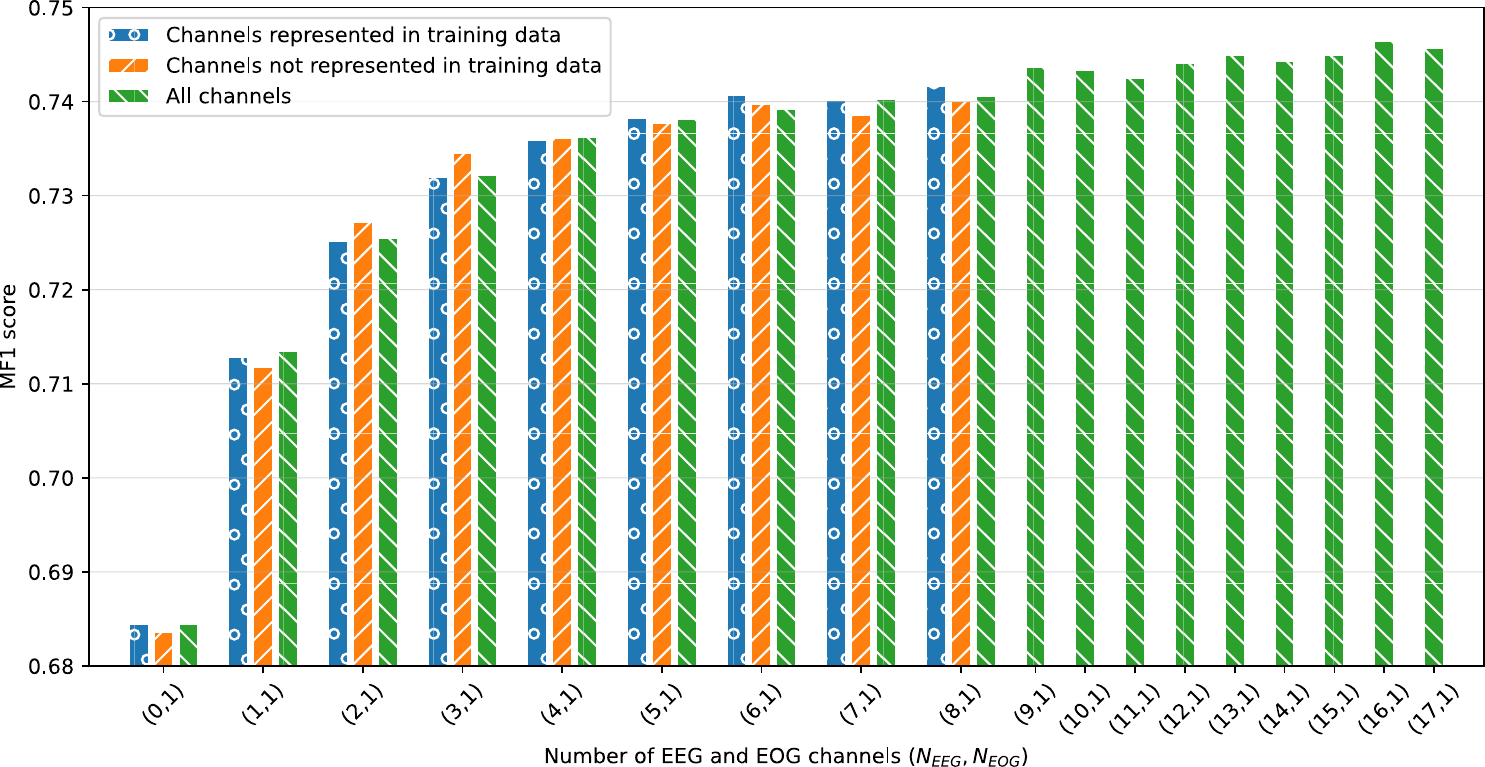}
  \caption{
    Generalization of AnySleep to unseen channels across varying channel counts.
    AnySleep was evaluated on the mass-c1 and mass-c3 datasets with one fixed EOG channel and three EEG configurations: (i) only EEG derivations used during training (blue), (ii) only EEG channels not used during training (orange), and (iii) all available EEG channels (green).
    Performance is reported as recording-wise macro F1 scores and grouped by the combination of EEG ($N_\text{EEG}$) and EOG ($N_\text{EOG}$) channels.
    For each channel-count condition, we randomly sampled 5000 recordings with replacement and evaluated each on a randomly chosen subset of channels of that size (without replacement).
    Evaluations were repeated for three independent training runs, and we report the average scores over recordings and runs.
    Model performance increased with the number of channels across all conditions, with no substantial difference between seen and unseen channels, indicating that AnySleep generalizes to previously unseen channels without performance degradation.
  }
  \label{sup-fig:nchannels-mass}
\end{figure}

\begin{table}[h]
  \centering
  \small
  \begin{tabular}{lr|ccccccc}
    Model     & $N_{\text{Rec}}$ & Wake        & N1          & N2          & N3          & REM         & MF1         & Coh. $\kappa$ \\
    \toprule
    AnySleep  & 28               & 0.91 (.002) & 0.46 (.009) & 0.90 (.000) & 0.91 (.005) & 0.88 (.002) & 0.81 (.002) & 0.82 (.002)   \\
    \USleep{} & 28               & 0.89 (.010) & 0.47 (.001) & 0.89 (.005) & 0.87 (.019) & 0.87 (.002) & 0.80 (.004) & 0.79 (.006)   \\
  \end{tabular}
  \caption{
    Performance of AnySleep and \USleep{} on the high-density held-out ANPHY~\cite{Wei2024} dataset.
    This dataset contains 28 recordings from 13 female and 15 male healthy subjects (mean age 32.46 $\pm$ 6.10 years).
    Scores were calculated using the 83 EEG and 2 EOG channels contained in each recording (see Supplementary Table~\ref{sup-tab:channels}).
    For \USleep{}'s majority voting approach, this required 166 evaluations of different EEG-EOG channel pairs.
    The models were trained three times with different random seeds, and we report the mean scores and standard deviations (in parentheses).
  }
  \label{sup-tab:high-density}
\end{table}

\begin{figure}[h]
  \centering
  \includegraphics[width=0.7\linewidth]{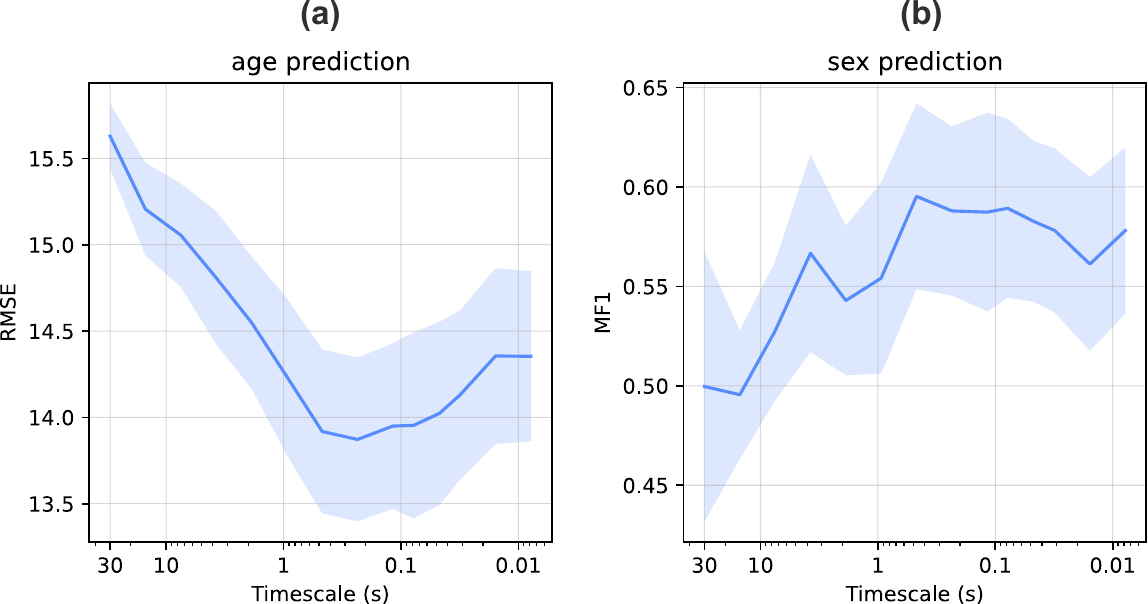}
  \caption{
    Prediction of physiological characteristics (age, sex) from triplet features derived from AnySleep's high-frequency sleep stages at different timescales on held-out datasets.
    \textbf{(a)} Root mean squared error (RMSE) of a random forest (RF) regression model predicting age from triplet features for 116 subjects from ISRUC sg1--3; recording-wise age predictions were calculated as the average of predictions across 1.5~h blocks (see Section~\ref{ssec:high-freq}).
    \textbf{(b)} Macro F1 (MF1) scores of an RF classifier predicting sex (male vs female) for 118 subjects from ISRUC sg1--3.
    For each timescale and task in (a) and (b), 50 RF models were trained on features derived from each of three independent AnySleep training runs (150 RF models in total); the blue line and blue shaded area show mean score and standard deviation across these models.
    Model performance improved as the timescale of the high-frequency sleep stages decreased, suggesting that AnySleep's high-frequency sleep stages contain additional age- and sex-associated information compared to 30-s sleep stages.
  }
  \label{sup-fig:high-freq-age-sex}
\end{figure}

\begin{figure}[h]
  \centering
  \includegraphics[width=\linewidth]{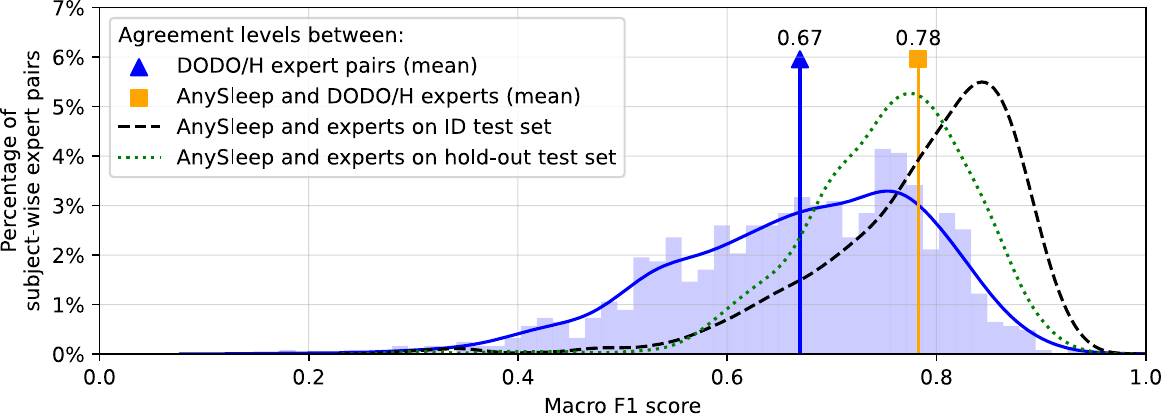}
  \caption{
    Comparison between AnySleep-expert and expert-expert agreement on held-out DODO and DODH datasets.
    Blue bars show a histogram of agreements levels (measured by MF1 scores) between pairs of experts for each subject in the combined DODO/H datasets, which were annotated by six different experts.
    We also show a density estimate of the expert agreement distribution (blue curve) and the average expert-expert agreement level (blue vertical line).
    AnySleep-expert agreement was calculated subject-wise for each available expert annotation and aggregated across the three training runs, with the average agreement level shown in orange.
    For context, density estimates show the distributions of AnySleep-expert agreement across all in-distribution test datasets (black dashed curve) and hold-out test datasets (green dotted curve).
    AnySleep-expert agreement on DODO/DODH was comparable to or higher than typical expert-expert agreement, indicating that the model performs within the range of human inter-rater variability on these datasets.
  }
  \label{sup-fig:inter-rater}
\end{figure}

\begin{figure}[h]
  \centering
  \includegraphics[width=\linewidth]{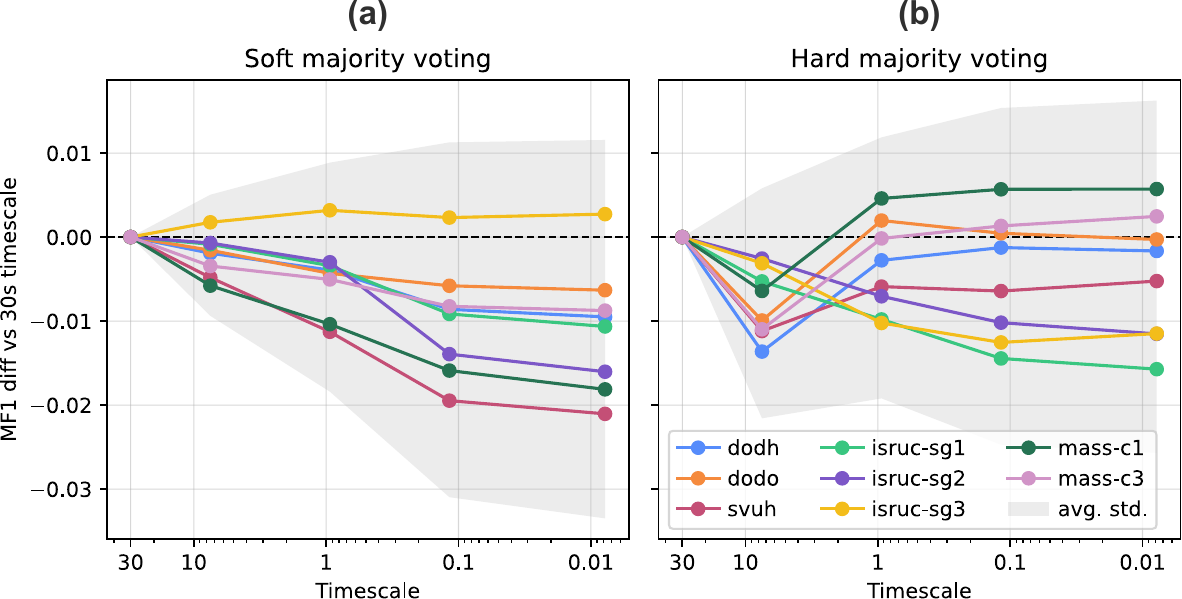}
  \caption{
    Comparison of approaches for aggregating AnySleep's high-frequency representations into 30-s labels on the hold-out test datasets.
    Each curve plots the difference in macro F1 (MF1) between AnySleep evaluated at a given timescale and evaluations at the 30-s baseline.
    At the 30-s timescale, internal high-frequency representations are aggregated with an average pooling layer in the segment classifier (see Figure~\ref{fig:architecture}c).
    At lower timescales, the internal high-frequency representations are first downsampled into the respective timescale by the average pooling layer, passed through the remainder of the segment classifier, and then aggregated into 30-s labels using either soft majority voting (panel \textbf{(a)}; average of output logits over the sub-intervals of each 30-s window) or hard majority voting (panel \textbf{(b)}; majority count of predicted sleep stages with ties broken at random).
    MF1 differences were calculated for each subject and each of three training runs.
    We report the average MF1 difference over subjects and training runs for each hold-out test dataset, as well as the average standard deviation (avg. std.) across datasets.
    Results varied by dataset, as soft majority voting largely reduced MF1 except for isruc-sg3 while hard majority voting improved MF1 for some datasets such as mass-c1 but not for others.
  }
  \label{sup-fig:high-freq-agg}
\end{figure}

\end{document}